\documentclass{article} 
\usepackage{iclr2027_conference,times}

\usepackage{amsmath,amsfonts,bm}

\def\eqref#1{equation~\ref{#1}}

\def\1{\bm{1}}

\DeclareMathAlphabet{\mathsfit}{\encodingdefault}{\sfdefault}{m}{sl}
\SetMathAlphabet{\mathsfit}{bold}{\encodingdefault}{\sfdefault}{bx}{n}

\def\gA{{\mathcal{A}}}

\def\gX{{\mathcal{X}}}
\def\gY{{\mathcal{Y}}}

\newcommand{\E}{\mathbb{E}}
\newcommand{\Ls}{\mathcal{L}}

\newcommand{\sigmoid}{\sigma}

\newcommand{\KL}{D_{\mathrm{KL}}}

\newcommand{\Cov}{\mathrm{Cov}}

\usepackage{amsmath}
\usepackage{microtype}
\usepackage{amssymb}
\usepackage{float}
\usepackage{placeins}
\usepackage{graphicx}
\usepackage{booktabs}
\usepackage{tabularx}
\usepackage{array}
\usepackage{colortbl}
\usepackage{xcolor}
\usepackage{xspace}
\usepackage{hyperref}
\usepackage{url}

\usepackage{tikz}

\definecolor{PIShapeRoseDark}{HTML}{9E416C}    
\definecolor{PIShapeBlush}{HTML}{F7E7ED}       
\definecolor{PIShapeBlueSoft}{HTML}{E8EDF5}    
\definecolor{PIShapeGroupGrey}{HTML}{F3F0F2}   
\definecolor{PIShapeCharcoal}{HTML}{3F3F3F}    
\definecolor{PIShapeZero}{HTML}{707070}        

\definecolor{PIHeatPositive}{HTML}{5F947E}
\definecolor{PIHeatNegative}{HTML}{BC758B}
\newcommand{\effectshade}[2]{%
  \pgfmathsetmacro{\PIHeatMix}{80*min(1,abs(#1)/(#2))}%
  \ifdim #1pt<0pt\relax
    \edef\PIHeatFill{PIHeatNegative!\PIHeatMix!white}%
  \else
    \edef\PIHeatFill{PIHeatPositive!\PIHeatMix!white}%
  \fi
  \expandafter\cellcolor\expandafter{\PIHeatFill}%
}
\newcolumntype{E}{>{\columncolor{white}[0pt][0pt]}r}
\newcolumntype{G}{>{\columncolor{white}[0pt][0pt]}c}

\definecolor{PIViewAO}{HTML}{D783A3}
\definecolor{PIViewGI}{HTML}{C36993}
\definecolor{PIViewKP}{HTML}{AF4F83}
\definecolor{PIViewCS}{HTML}{96447D}
\definecolor{PIViewSU}{HTML}{7D3876}
\definecolor{PIViewFT}{HTML}{642D70}

\newcommand{\viewsq}[1]{{\color{#1}\rule[0.06ex]{0.9ex}{0.9ex}}\kern0.35em}

\newcommand{\hdrl}[2]{\begin{tabular}[b]{@{}l@{}}\textbf{#1}\\\textbf{#2}\end{tabular}}

\newcommand{\hdrr}[2]{\begin{tabular}[b]{@{}r@{}}\textbf{#1}\\\textbf{#2}\end{tabular}}

\newcommand{\tablefont}{\fontsize{8}{9.6}\selectfont}
\newcommand{\tablesetup}{\tablefont\setlength{\tabcolsep}{4pt}\renewcommand{\arraystretch}{1.12}}

\newcommand{\ci}[3]{{$#1_{[#2,#3]}$}}

\definecolor{PIShapeRoseDark}{HTML}{9E416C}
\definecolor{PIShapeBlush}{HTML}{F7E7ED}
\definecolor{PIShapeBlueSoft}{HTML}{E8EDF5}
\hypersetup{
  colorlinks=true,
  linkcolor=PIShapeRoseDark,
  citecolor=PIShapeRoseDark,
  urlcolor=PIShapeRoseDark
}

\newcommand{\DR}{direct-response\xspace}

\newcommand{\AO}{\textsc{Answer Only}\xspace}
\newcommand{\GI}{\textsc{Gist}\xspace}
\newcommand{\KP}{\textsc{Key Points}\xspace}
\newcommand{\CS}{\textsc{Clean Solution}\xspace}
\newcommand{\SU}{\textsc{Summary}\xspace}
\newcommand{\FT}{\textsc{Full Trace}\xspace}
\newcommand{\codeurl}{https://github.com/xiuyuz/opsd-reference-study}
\newcommand{\dataseturl}{https://huggingface.co/datasets/xiuyuz/ample-math}
\newcommand{\coderepo}{\href{\codeurl}{GitHub}}
\newcommand{\datasetrepo}{\href{\dataseturl}{Hugging Face}}
\newcommand{\resourcelinks}{%
  \href{\codeurl}{%
    \raisebox{-0.22em}{\includegraphics[height=1.35em]{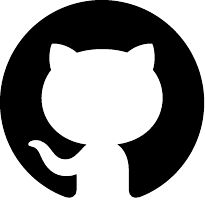}}%
    \hspace{0.4em}Code}%
  \hspace{2em}%
  \href{\dataseturl}{%
    \raisebox{-0.22em}{\includegraphics[height=1.35em]{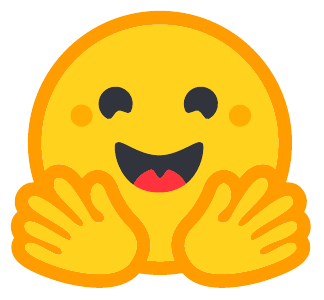}}%
    \hspace{0.4em}AMPLE-Math}%
}

\title{What Does Privileged Information Add to\\
On-Policy Self-Distillation?}

\author{XiuYu Zhang, Wei Chow, Junfeng Fang, Xingyu Zhu, Zhenkai Liang, Tat-Seng Chua\\
\normalfont National University of Singapore}

\iclrfinalcopy
\begin{document}

\maketitle
\vspace{-1.8\baselineskip}
\noindent\hspace*{\tabcolsep}\resourcelinks\par
\vspace{0.8\baselineskip}

\begin{abstract}
On-policy self-distillation (OPSD) lets a language model learn from a frozen copy of itself that sees an answer or a worked solution.
Giving the teacher this extra information seems to offer the student more to learn, but how much does it add beyond distillation itself?
To isolate that contribution, we construct \textsc{AMPLE-Math}, a reusable suite of 5,319 mathematical problems with six reasoning views that share the same answer, and compare each view with matched reference-free distillation.
With a thinking-enabled teacher supervising direct-response rollouts, reference-free distillation accounts for much of Qwen3-1.7B's improvement under thinking-enabled evaluation, both in domain and on external benchmarks.
Evidence for an additional reference benefit is modest in Qwen, strongest for a polished solution, whereas complete traces add two percentage points in SmolLM3-3B at step 50.
These benefits depend on the student being trained. At the same checkpoint, replacing short direct-response rollouts with long thinking-enabled rollouts turns gains into losses in both families while the problems, references, and evaluation stay fixed.
Teacher profiles and matched loss interventions in Qwen further show that changing token-level supervision can leave student behavior largely unchanged.
Together, these findings suggest that OPSD can improve access to existing reasoning capabilities through parameters shared by direct-response and thinking-enabled inference.
The value of a privileged reference is what it adds to this cross-mode transfer, not how much of the solution it reveals.
\end{abstract}

\section{Introduction}
\label{sec:introduction}

A worked solution gives a teacher information that its student does not have.
On-policy self-distillation (OPSD) uses this asymmetry to improve a language model without a separate, larger teacher.
A frozen copy of the model sees the solution and scores responses generated by the student, which sees only the problem \citep{agarwal2024onpolicy,zhao2026opsd}.
As in learning with privileged information(PI) \citep{vapnik2009new,lopezpaz2016unifying}, the extra information supports training without being needed at inference.
Similarly, the best content and form of privileged information for this type of training remains undecided.
A fuller reference reveals more about the solution, but does it give the student more to learn?

Improvement over the base model alone does not answer that question.
Recent studies find useful supervision with absent or other-problem references \citep{shrestha2026rethinking,ichihara2026op2sd}, while structured guidance can outperform complete solutions \citep{zhao2026morepi}.
These findings establish the need to separate what a reference contributes from what distillation already achieves.
We build on these controls by holding the answer fixed across reasoning representations and tracking their effects as the student changes.
In OPSD, the teacher scores the student's own prefixes rather than supplying a solution to reproduce. A reference must therefore help the teacher give useful feedback on the student's attempt.

\begin{figure}[!t]
\centering
\includegraphics[width=\linewidth]{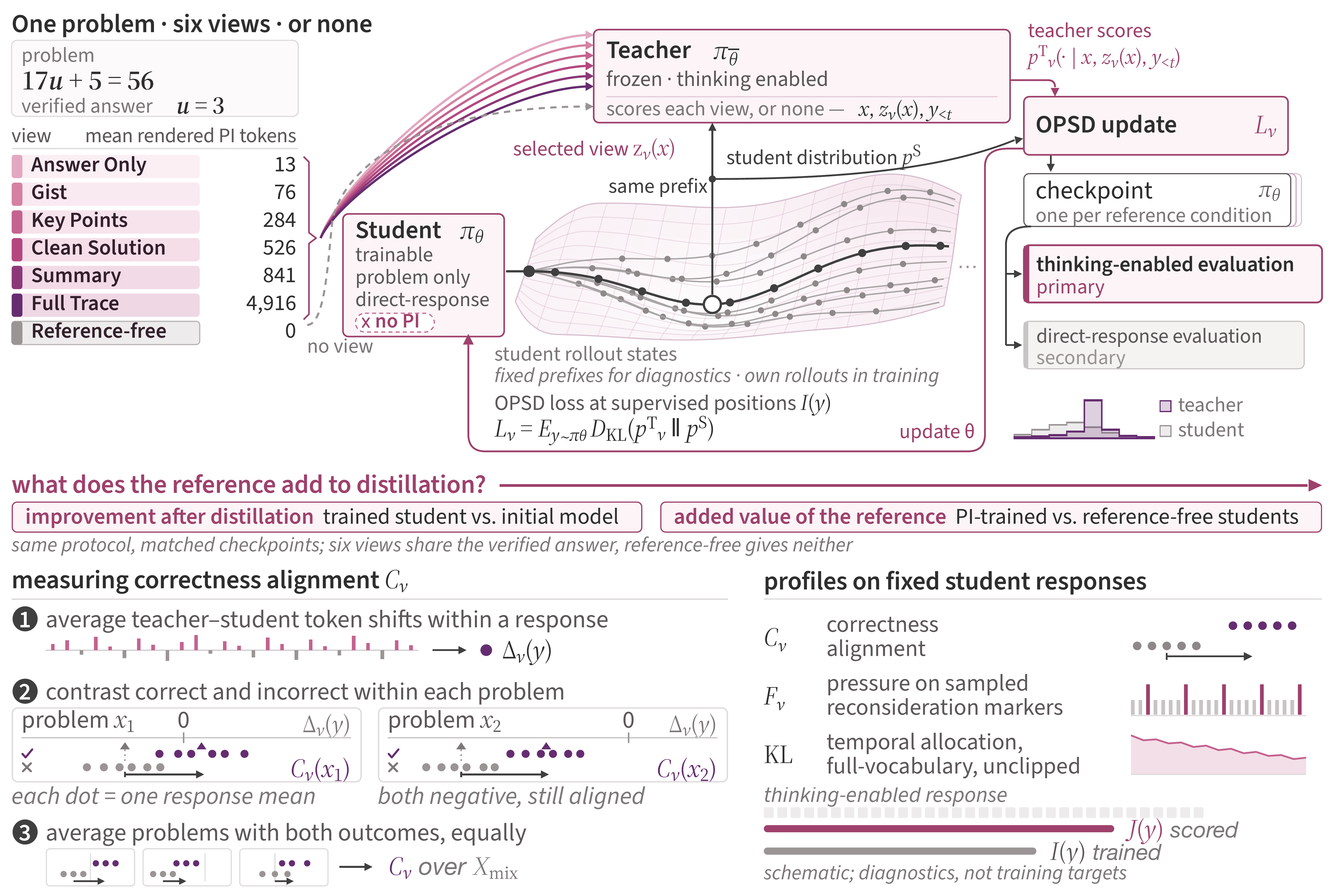}
\caption{\textbf{What does the reference add to distillation?}
The frozen, thinking-enabled teacher scores student prefixes with or without privileged information.
Lower panels illustrate correctness alignment and diagnostic profiles.
The loss is schematic, with clipping specified in Equation~\ref{eq:opsd}.}
\label{fig:transfer}
\end{figure}

To study this interaction while holding answer information fixed, we construct \textsc{AMPLE-Math} (Answer-Matched Privileged Levels of Explanation) from OpenThoughts-114k \citep{guha2025openthoughts}.
Each of its 5,319 problems has six views, from \AO to \FT, sharing one verified answer while varying the reasoning representation.
We compare each view with a reference-free control under the same training configuration (Figure~\ref{fig:transfer}).
A thinking-enabled teacher supervises direct-response rollouts \citep{zhao2026opsd,ichihara2026op2sd}, and we evaluate the student with thinking enabled.
Comparisons in Qwen3-1.7B~\citep{qwen2025qwen3} and SmolLM3-3B~\citep{huggingface2025smollm3} follow the reference's contribution across checkpoints.

Removing the reference does not fully remove the teacher--student asymmetry.
The original OPSD configuration and several subsequent methods pair a
thinking-enabled teacher with a direct-response student~\citep{zhao2026opsd,lin2026renio,kara2026feedback,li2026uopsd}.
As a result, the thinking-enabled teacher still scores prefixes produced with thinking disabled even without any PI~\citep{ichihara2026op2sd,shrestha2026rethinking}.
Learning from these scores changes parameters shared by both student inference modes, offering a route to stronger thinking-enabled reasoning without additional solution information.
Much of Qwen's improvement is already present without a reference, with modest evidence of an additional benefit from \CS.
In SmolLM3, \FT adds two points beyond reference-free training at step 50.
The same references nevertheless accompany gains under direct-response training rollouts and losses under long thinking-enabled rollouts in both families.
What the reference contributes therefore depends on the student responses it helps the teacher supervise.

To understand this dependence, we examine what the teacher's scores encourage on student responses.
Prior work points to preferences for the supplied solution \citep{harne2026biased}, likelihood shifts on student tokens \citep{nguyen2026likelihood}, and suppressed reconsideration \citep{kaur2026rethinking} as possible sources of poor transfer.
Our Qwen profiles show that correctness alignment changes with the responses being scored, while sampled correction markers receive negative pressure in both prefix modes.
Relaxing loss at these markers barely changes their use, and an apparent benefit from broader supervision mostly disappears when checkpoints are matched.
Teacher profiles identify explanations to test, while matched training outcomes determine which changes help.

\noindent\textbf{Our contributions can be summarized as follows.}
First, we construct \textsc{AMPLE-Math}, a reusable supervision suite for isolating reasoning representation through six answer-matched views per problem, with audits and frozen splits.
Second, controlled comparisons in two model families separate the benefits of cross-mode distillation from a reference's added value and show that changing student training trajectories can reverse transfer.
Third, teacher profiles and matched interventions distinguish changes to supervision from changes in student behavior, and show that stopping time explains most of an apparent gain from broader loss coverage.

\section{A Controlled Study of Privileged Supervision}
\label{sec:controlled-study}

Our study separates the contribution of a privileged reference from the gains of distillation, using thinking-enabled student performance as the primary outcome.

\subsection{OPSD on Student-Generated Prefixes}
\label{sec:opsd-setup}

For a problem $x$, view $v$ supplies PI $z_v(x)$ to a frozen teacher $\pi_{\bar\theta}$.
The student $\pi_\theta$ starts from the same backbone and generates $y=(y_1,\ldots,y_L)$ from the problem alone under rollout configuration $m_S$.
For a vocabulary token $a$ at prefix $s_t=(x,y_{<t})$, the student's next-token distribution is \mbox{$p^S_t(a)=\pi_\theta(a\mid x,y_{<t};m_S)$}.
The teacher scores the same prefix using \mbox{$p^T_{v,t}(a)=\pi_{\bar\theta}(a\mid x,z_v(x),y_{<t};m_T,\tau)$}.
Here $m_T$ specifies the teacher's reasoning configuration and $\tau$ its scoring temperature.
The reference-free control omits the PI section but retains the thinking-enabled teacher scoring direct-response student prefixes.
It is therefore cross-mode self-distillation, not a comparison of identical teacher and student distributions.
For the supervised positions $I(y)$, the OPSD loss is
\begin{equation}
 \Ls_v(\theta;m_S)=
 \E_{x,\,y\sim\pi_\theta(\cdot\mid x;m_S)}
 \left[\frac{1}{|I(y)|}\sum_{t\in I(y)}
 D_{\mathrm{gKL}}\!\left(p^T_{v,t}\,\|\,p^S_t\right)\right],
 \label{eq:opsd}
\end{equation}
where $D_{\mathrm{gKL}}$ clips each vocabulary term in the teacher-to-student forward KL from above at $0.05$ before summation \citep{zhao2026opsd}.
The teacher distribution and sampled completion are held fixed when differentiating the loss.
With the same prompt, thinking mode, initial parameters, and scoring temperature, a reference-free teacher would match the student and give zero initial loss.
PI then supplies the initial discrepancy, while later updates also separate the student's parameters from the frozen teacher.

\subsection{What a Privilege Profile Measures}
\label{sec:profile}

Before training, we profile Qwen3-1.7B's token-level supervision on 512 problems, using four unprivileged completions per problem under each rollout configuration.
For each completion $y$, let $J(y)$ index the longest prefix fitting every teacher context, capped at 1,024 completion tokens for direct-response profiles.
Scoring the same tokens under every view gives the mean log-probability shift
\begin{equation}
 \Delta_v(y)=\frac{1}{|J(y)|}\sum_{t\in J(y)}\big[\log p^T_{v,t}(y_t)-\log p^S_t(y_t)\big].
 \label{eq:privileged-shift}
\end{equation}

Let $\gY_x^+$ and $\gY_x^-$ contain the correct and incorrect profiled samples for problem $x$, and let $\gX_{\mathrm{mix}}$ contain problems with both outcomes.
\textbf{Correctness alignment} $C_v$ contrasts responses for each $x\in\gX_{\mathrm{mix}}$, then averages these problems equally:
\begin{equation}
 C_v(x)=\frac{\sum_{y\in\gY_x^+}\Delta_v(y)}{|\gY_x^+|}-\frac{\sum_{y\in\gY_x^-}\Delta_v(y)}{|\gY_x^-|},\qquad
 C_v=\frac{1}{|\gX_{\mathrm{mix}}|}\sum_{x\in\gX_{\mathrm{mix}}}C_v(x).
 \label{eq:correctness-alignment}
\end{equation}
Positive alignment means larger teacher--student shifts for correct than incorrect responses, on average within problems.
\textbf{Correction pressure} $F_v$ averages the shift at sampled reconsideration markers such as \emph{wait}, \emph{but}, and \emph{check}, with negative values indicating lower marker probability.
\textbf{Temporal KL allocation} is the share of full-vocabulary $\KL(p^T_{v,t}\,\|\,p^S_t)$ in each quarter of the retained span.
We use unclipped KL and normalize quarters within this span, not the training loss support.

Holding student responses fixed lets us compare how different references
shape supervision on the same prefixes.
Correctness alignment captures whether correct responses receive more
favorable shifts than incorrect ones, while marker pressure and KL
allocation guide our tests of marker weighting and loss coverage.
We then measure transfer after training and repeat the profiles on trained
students' responses to examine which supervision patterns persist as the
student learns.
Appendix~\ref{app:theory} derives the local reweighting interpretation
of correctness alignment.

\subsection{AMPLE-Math: Answer-Matched Privileged Views}
\label{sec:pi-shape-math}

We construct \textsc{AMPLE-Math} by pairing 5,319 mathematical problems from OpenThoughts-114k \citep{guha2025openthoughts} with six answer-matched views.
\AO supplies no reasoning body, \GI the central method, and \KP ordered steps.
\CS supplies a polished solution, \SU a narrative summary, and \FT the complete source trace.
Every view ends with the same canonical answer section.
Mean rendered lengths range from 13 to 4,916 Qwen3 tokens, describing reasoning density rather than assumed quality.
\mbox{\GI, \KP, and \SU} are generated from the intact source trace by Qwen3.6-35B-A3B-FP8 \citep{qwen2026qwen36} and checked for source fidelity by the same model under a separate review prompt.
\AO is deterministic, while \CS and \FT are source-derived.

We annotate problem difficulty using 8--16 problem-only samples per question from each of three evaluator models and a binomial Rasch model \citep{rasch1960probabilistic}.
Separately, four unprivileged direct-response samples per problem from the frozen Qwen3-1.7B base define three model-relative success bands for the experimental splits: three or four correct, one or two correct, and none correct.
The zero-success band requires a correct generation from a structured-PI teacher view.
Equal band quotas give train, development, and test sets of 1,536, 192, and 384 problems.
SmolLM3 uses the same development and test problems but a training split rebuilt from its own direct-response outcomes.
Appendix~\ref{app:dataset} details construction, filtering, fidelity audits, difficulty estimation, and splits.

\subsection{Training and Evaluation}
\label{sec:study-design}

\noindent\textbf{Models and rollouts.} The main experiments train Qwen3-1.7B with LoRA adapters \citep{hu2022lora} ($r=64$, $\alpha=128$) for 100 optimizer steps, with SmolLM3-3B providing the cross-family test \citep{qwen2025qwen3,huggingface2025smollm3}.
The teacher is always the frozen, thinking-enabled backbone.
Our primary configuration uses \DR training followed by thinking-enabled evaluation, an asymmetry also examined in other OPSD studies \citep{zhao2026opsd,ichihara2026op2sd,shrestha2026rethinking}.
The student generates at most 1,024 tokens with thinking disabled, while the thinking-enabled training comparison extends rollout generation.

\noindent\textbf{Evaluation and statistics.} Primary evaluation uses four unprivileged thinking-enabled samples per problem.
Secondary comparisons evaluate the same checkpoints with thinking disabled.
Avg@4 averages the fraction of correct samples across problems.
In-domain evaluation uses a 16,384-token budget and regenerates capped responses with the same sampling seed and up to 32,512 tokens.
The length analyses grade the first 4,096, 8,192, and 16,384 saved token IDs from the initial pass, rather than generating new responses at those limits.
External evaluation uses 12 thinking-enabled samples per problem \citep{zhao2026opsd} on AIME 2024 and 2025 \citep{maa2024aime,maa2025aime} and HMMT February 2025 \citep{hmmt2025february}.

The principal reference comparisons use steps 50 and 100 in both families.
Paired 95\% intervals use 10,000 problem-cluster bootstrap resamples.
For the initial six-view control tests, we compare seed-0 students with three-seed controls and apply Holm correction across the six contrasts.
Follow-up seed averages are labeled in the figures and tables. Their intervals resample problems after averaging the observed seeds, with seed-level $t$ intervals reported as a separate check.
Full profiling, training, and evaluation details appear in Appendix~\ref{app:experimental-details}.

\section{What Privileged References Add}
\label{sec:trajectory-relative-transfer}

Improvement over the base model can come from distillation itself, so we compare students trained with and without a teacher reference to identify what privileged information adds.

\subsection{Distillation Gains and Reference Contributions}
\label{sec:six-view-transfer}

\begin{figure}[!t]
\centering
\includegraphics[width=\linewidth]{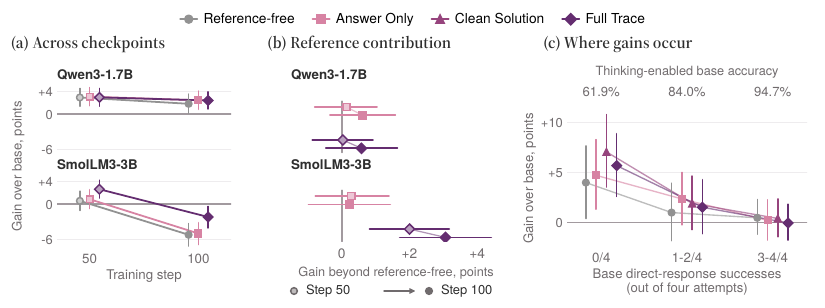}
\caption{\textbf{Distillation gains and reference contributions are different quantities.}
Direct-response training and thinking-enabled evaluation throughout.
(a) Gains over the base across checkpoints. (b) Paired gains beyond reference-free training.
(c) Qwen step-100 gains grouped by base direct-response successes. Top annotations give thinking-enabled base accuracy.
Error bars show unadjusted 95\% problem-cluster bootstrap confidence intervals (384 problems, 128 per subgroup).
Three seeds per configuration, except four for Qwen \AO/\FT.}
\label{fig:sixview}
\end{figure}

In Qwen, much of the improvement attributed to privileged supervision is already present without a reference (Figure~\ref{fig:sixview}a, Table~\ref{tab:profile-prediction}).
The reference-free student improves both in domain and on external benchmarks, and even the wrong-answer control improves over the base.
More reasoning brings no consistent increase in gains.
At step 100, \AO and \FT sit about $0.6$ points above the reference-free student, with both intervals including zero.

\CS gives modest evidence that a reference can add value in Qwen.
Across three seeds per configuration, its step-100 advantage over reference-free training is $1.30\,[0.20,2.41]$ points.
The interval excludes zero before adjustment, but the effect does not survive Holm correction across six views.
Thus the evidence favors a small, view-specific contribution rather than a general benefit from supplying more of the solution.

Gains are largest on problems the base never solved in four direct-response attempts, but solves $62\%$ of the time with thinking enabled (Figure~\ref{fig:sixview}c).
This pattern holds with and without PI, and \CS's additional benefit is also largest in this group.
In the separate original-data validation, external gains concentrate on problems solved in some but not all of the base's twelve thinking-enabled attempts.
Both patterns fit the shared-parameter interpretation.
Learning from thinking-enabled teacher scores on direct-response prefixes may also improve the student's thinking-enabled reasoning.

\begin{table}[!t]
\caption{\textbf{Six-view transfer in domain and on external benchmarks.}
Qwen3-1.7B gains over each protocol's frozen base after direct-response training, averaged over the listed training seeds (in/ex denotes in-domain/external).
Wrong answer uses \AO with another problem's answer.
All evaluation is thinking-enabled, with 30 problems per external benchmark.
Subscripts give 95\% problem-cluster bootstrap intervals.
Green/red shading shows positive/negative gains on a shared $\pm10$-point scale.}
\label{tab:profile-prediction}
\centering
\tablesetup
\setlength{\tabcolsep}{1.25pt}
\renewcommand{\ci}[3]{{$#1_{\scalebox{0.8}{$\scriptstyle[#2,#3]$}}$}}
\begin{tabular*}{\linewidth}{@{\extracolsep{\fill}}lrr*{5}{E}@{}}
\toprule
& & & \multicolumn{1}{c}{\textbf{In domain}} & \multicolumn{4}{c}{\textbf{External, $\Delta$ Avg@12 at step 50}} \\
\cmidrule(lr){4-4}\cmidrule(l){5-8}
\hdrl{}{Reference} &
\hdrr{PI}{tokens} &
\hdrr{Seeds}{(in/ex)} &
\hdrr{Step 100}{$\Delta$ Avg@4} &
\hdrr{}{Pooled} &
\hdrr{AIME}{2024} &
\hdrr{AIME}{2025} &
\hdrr{HMMT}{2025} \\
\midrule
Reference-free & 0 & 3/3 & \effectshade{+1.80}{10}\ci{+1.80}{0.20}{3.47} & \effectshade{+4.07}{10}\ci{+4.07}{1.73}{6.54} & \effectshade{+8.24}{10}\ci{+8.24}{3.61}{13.15} & \effectshade{+1.67}{10}\ci{+1.67}{-1.85}{5.37} & \effectshade{+2.31}{10}\ci{+2.31}{-1.39}{6.11} \\
Wrong answer & 13 & 3/3 & \effectshade{+1.67}{10}\ci{+1.67}{0.07}{3.32} & \effectshade{+4.14}{10}\ci{+4.14}{1.33}{7.07} & \effectshade{+6.30}{10}\ci{+6.30}{0.28}{12.59} & \effectshade{+2.87}{10}\ci{+2.87}{-1.11}{6.94} & \effectshade{+3.24}{10}\ci{+3.24}{-1.02}{7.50} \\
\midrule
\viewsq{PIViewAO}\AO & 13 & 4/4 & \effectshade{+2.41}{10}\ci{+2.41}{0.86}{3.99} & \effectshade{+3.73}{10}\ci{+3.73}{1.27}{6.41} & \effectshade{+8.06}{10}\ci{+8.06}{3.61}{13.06} & \effectshade{+1.53}{10}\ci{+1.53}{-1.18}{4.44} & \effectshade{+1.60}{10}\ci{+1.60}{-3.06}{7.01} \\
\viewsq{PIViewGI}\GI & 76 & 1/1 & \effectshade{+2.15}{10}\ci{+2.15}{0.26}{4.04} & \effectshade{+4.35}{10}\ci{+4.35}{1.30}{7.59} & \effectshade{+6.67}{10}\ci{+6.67}{0.56}{13.06} & \effectshade{+3.89}{10}\ci{+3.89}{-0.56}{8.33} & \effectshade{+2.50}{10}\ci{+2.50}{-3.06}{8.06} \\
\viewsq{PIViewKP}\KP & 284 & 1/1 & \effectshade{+2.80}{10}\ci{+2.80}{0.72}{4.88} & \effectshade{+3.61}{10}\ci{+3.61}{0.37}{7.04} & \effectshade{+3.89}{10}\ci{+3.89}{-2.78}{10.56} & \effectshade{+5.00}{10}\ci{+5.00}{0.28}{9.72} & \effectshade{+1.94}{10}\ci{+1.94}{-3.33}{7.78} \\
\viewsq{PIViewCS}\CS & 526 & 3/1 & \effectshade{+3.10}{10}\ci{+3.10}{1.54}{4.73} & \effectshade{+4.91}{10}\ci{+4.91}{2.04}{7.96} & \effectshade{+7.78}{10}\ci{+7.78}{2.22}{13.89} & \effectshade{+4.44}{10}\ci{+4.44}{0.83}{8.89} & \effectshade{+2.50}{10}\ci{+2.50}{-2.50}{8.06} \\
\viewsq{PIViewSU}\SU & 841 & 1/1 & \effectshade{+2.54}{10}\ci{+2.54}{0.71}{4.43} & \effectshade{+3.61}{10}\ci{+3.61}{0.83}{6.48} & \effectshade{+6.94}{10}\ci{+6.94}{1.39}{12.78} & \effectshade{+1.67}{10}\ci{+1.67}{-2.50}{6.11} & \effectshade{+2.22}{10}\ci{+2.22}{-1.94}{6.67} \\
\viewsq{PIViewFT}\FT & 4,916 & 4/4 & \effectshade{+2.38}{10}\ci{+2.38}{0.91}{3.86} & \effectshade{+2.29}{10}\ci{+2.29}{-0.16}{4.88} & \effectshade{+5.97}{10}\ci{+5.97}{1.39}{10.97} & \effectshade{+0.49}{10}\ci{+0.49}{-3.47}{4.65} & \effectshade{+0.42}{10}\ci{+0.42}{-3.12}{4.24} \\
\bottomrule
\end{tabular*}
\end{table}

The reference makes a clearer contribution in SmolLM3, where \FT adds $2.0$ points over reference-free training at step 50 with thinking enabled (Figure~\ref{fig:sixview}b).
\label{sec:evaluation-dependent-reference}
By step 100 all three configurations have fallen below the base, but \FT loses less than the reference-free and \AO students.
The same reference helps the earlier student and softens the later deterioration, two different contributions that a gain over the base alone would not distinguish.

The same students rank references differently when answering directly.
\label{sec:response-prefixes}
At step 100, Qwen's \AO and reference-free students exceed \FT by $5.32$ and $10.81$ points under direct-response evaluation.
SmolLM3's step-50 \FT advantage likewise changes from $+2.00$ points with thinking enabled to $-10.63$ when answering directly.
Qwen's direct-response winners also write longer responses, and grading only their first 4K tokens reverses the ordering.
The ranking changes accompany differences in response length, linking reference choice to how the student answers as well as how often it is correct.

\FloatBarrier
\subsection{Student Training Trajectories Change Transfer}
\label{sec:training-choices}
\label{sec:trajectory-reversal}

How the student answers also matters during training, when its responses determine the prefixes the teacher supervises.
To examine the thinking-model degradation reported by \citet{kaur2026rethinking}, we replace direct-response rollouts with thinking-enabled rollouts under \mbox{\AO}, \mbox{\CS}, and \mbox{\FT} supervision while keeping thinking-enabled evaluation fixed.
Thinking-enabled rollouts extend beyond the inherited \mbox{first-1,024-token} loss window, so the reasoning mode, the horizon, and the fraction of the response supervised change together.

\begin{figure}[htbp]
\centering
\includegraphics[width=\linewidth]{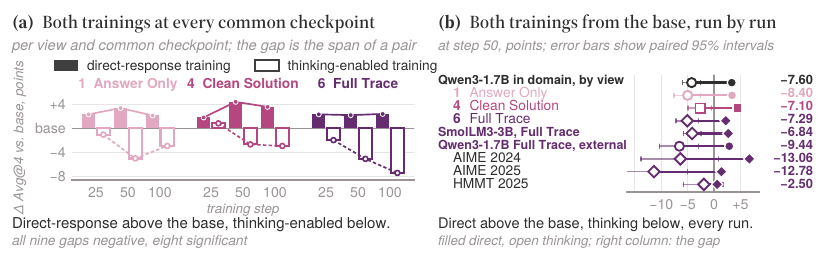}
\caption{\textbf{Changing training trajectories reverses transfer.} Thinking-enabled evaluation throughout.
(a) Avg@4 gains over the base across checkpoints.
(b) Step-50 gains: Qwen3-1.7B in domain, two-seed SmolLM3-3B under \FT, and Qwen under \FT externally.
Filled/open marks denote direct-response/thinking-enabled training.
Bold rows pool groups. Right-hand contrasts are thinking-enabled minus direct-response, with paired 95\% problem-cluster bootstrap intervals.}
\label{fig:rollout}
\end{figure}

\noindent\textbf{The same teacher and reference now accompany opposite transfer outcomes.}
At step 50 the thinking-enabled training configuration turns gains over the base into losses for all three views, with a penalty that persists across checkpoints (Figure~\ref{fig:rollout}a).
The step-50 \FT comparison also holds in SmolLM3 and on Qwen's external benchmarks (Figure~\ref{fig:rollout}b).
The reference is unchanged, but its supervision is now applied to a different kind of student attempt.

\section{How Student Responses Shape Supervision}
\label{sec:trajectory-relative-profiles}

To interpret the training-trajectory contrast, we profile probability gaps between privileged teachers and unprivileged students.
Teacher weights stay fixed, but predictions depend on the reference and the student's prefix.

\begin{figure}[H]
\centering
\includegraphics[width=\linewidth]{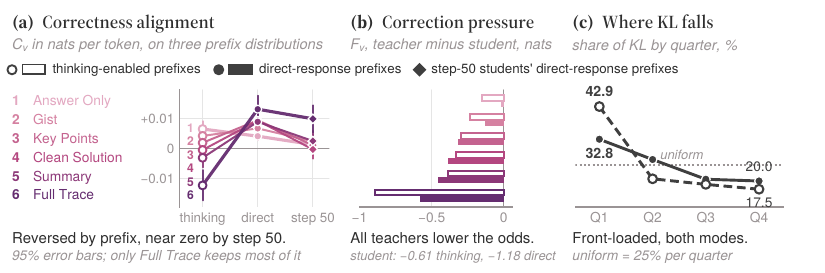}
\caption{\textbf{Teacher profiles separate correctness alignment, marker pressure, and supervision placement.}
(a) Correctness alignment $C_v$, (b) sampled reconsideration-marker pressure $F_v$, and (c) full-vocabulary KL shares by quarter of the scored span, not the training loss window.
Open/filled marks show frozen-base thinking-enabled/direct-response prefixes; diamonds show direct-response-trained students' prefixes at step 50.
Error bars show 95\% problem-cluster bootstrap intervals.}
\label{fig:profiles}
\end{figure}

\noindent\textbf{Correctness alignment changes with the responses being scored.}
\label{sec:correctness-alignment}
For \FT, correct responses receive larger average teacher--student log-probability shifts than incorrect ones on direct-response prefixes. The ordering reverses on thinking-enabled prefixes (Figure~\ref{fig:profiles}a).
On direct-response-trained students' responses, alignment is much smaller for five views, while \FT retains most of its initial value.
These profiles use new responses and updated student distributions.
Near-zero alignment means similar average shifts for correct and incorrect responses, not token-level agreement.

Denser references induce stronger overall shifts, which raw $C_v$ does not separate from correctness selectivity.
Restricting thinking-enabled profiles to their first 1,024 scored tokens also weakens differences across views.
Alignment thus describes supervision on a particular response distribution and span, not a fixed quality of the reference.

\noindent\textbf{Correction-marker suppression is shared across prefix modes.}
\label{sec:correction-pressure}
\citet{kaur2026rethinking} propose suppressed reconsideration as an explanation for thinking-model degradation.
In the frozen Qwen base, every view lowers sampled reconsideration-marker probability on average in both prefix modes (Figure~\ref{fig:profiles}b).
This shared sign does not distinguish the configurations with opposing transfer outcomes, motivating the targeted loss edits below.

\noindent\textbf{Diagnostic KL is concentrated near the response opening.}
\label{sec:front-loaded-pressure}
The first quarter of the retained span carries disproportionate diagnostic KL in both prefix modes (Figure~\ref{fig:profiles}c).
This locates teacher--student disagreement, not its association with correctness.
The inherited 1,024-token loss window \citep{zhao2026opsd} covers full direct-response rollouts but only the opening of long thinking-enabled rollouts.
We therefore test broader loss coverage for the long rollouts.

\section{Which Supervision Changes Matter for Transfer}
\label{sec:registered-stress-tests}

The profiles suggest where supervision might be changed, but only training reveals whether those changes help.
We test this in Qwen3-1.7B by modifying reference content and the distillation loss while keeping thinking-enabled evaluation fixed (Table~\ref{tab:mechanism-audit}).

\noindent\textbf{Reference-free gains do not make reference content irrelevant.}
\label{sec:surface-pathologies}
Replacing \CS or \KP with a length-matched view from another problem, including that problem's answer, lowers thinking-enabled accuracy by about two points relative to the genuine view in two single-seed comparisons.
Matching the reference's length does not preserve its effect when the text concerns a different problem.

Shortening a reference instead tests how much of a relevant solution is useful.
We cut \FT to each problem's \KP, \CS, or \SU length, retaining the trace opening but removing its later reasoning and canonical answer section.
These openings bring no clear gain with thinking enabled, although they improve direct-response accuracy over matched \FT by roughly $8$--$11$ points.
Compression therefore changes how the student benefits from the reference across inference modes.

\noindent\textbf{Relaxing correction-marker loss barely changes marker use.}
\label{sec:targeted-controls}
Under direct-response training, we exclude or downweight loss at sampled correction markers for \FT and \CS.\label{sec:correction-markers}
Neither edit produces an accuracy gain at step 100, and marker use remains almost unchanged.
Relaxing a penalty on sampled markers is distinct from encouraging the student to reconsider.

\begin{table}[!t]
\caption{\textbf{Matched interventions on reference content and the training signal.}
Seventeen single-seed comparisons, all evaluated with thinking enabled. Upper rows use direct-response training at step 100, lower rows compare thinking-enabled \FT loss windows at matched steps.
Avg@4 is in percent, accuracy changes in percentage points, and subscripts give 95\% problem-cluster bootstrap intervals.
$\Delta$ tokens gives mean length change, and $\Delta$ markers the relative change in mean per-response markers per 1,000 tokens, both versus the matched configuration.
Trace openings omit the canonical answer section.
Bold contrasts pass Holm's multiple-comparison correction within the two other-problem controls.
Colors follow Table~\ref{tab:profile-prediction}'s $\pm10$-point scale.}
\label{tab:mechanism-audit}
\centering
\tablesetup
\setlength{\tabcolsep}{1.9pt}
\begin{tabular}{l r EE rr}
\toprule
\hdrl{}{Intervention} &
\hdrr{}{Avg@4} &
\hdrr{$\Delta$ vs.}{matched} &
\hdrr{$\Delta$ vs.}{base} &
\hdrr{$\Delta$}{tokens} &
\hdrr{$\Delta$}{markers} \\
\midrule
\rowcolor{PIShapeGroupGrey}
\multicolumn{6}{l}{\textbf{Other-problem reference}\quad length-matched view, including the other problem's answer} \\
\CS other-problem view & \ci{81.84}{78.71}{84.83} & \effectshade{-1.95}{10}\ci{\bm{-1.95}}{-3.78}{-0.20} & \effectshade{+1.63}{10}\ci{+1.63}{-0.39}{3.65} & $-794$ & $+4.2\%$ \\
\KP other-problem view & \ci{80.86}{77.67}{83.85} & \effectshade{-2.15}{10}\ci{\bm{-2.15}}{-3.97}{-0.39} & \effectshade{+0.65}{10}\ci{+0.65}{-1.24}{2.48} & $-1{,}242$ & $+7.6\%$ \\
\rowcolor{PIShapeGroupGrey}
\multicolumn{6}{l}{\textbf{Trace opening}\quad \FT cut to the length of \CS, \KP, or \SU} \\
\FT $\to$ \CS & \ci{82.03}{78.97}{84.96} & \effectshade{-0.59}{10}\ci{-0.59}{-2.28}{1.11} & \effectshade{+1.82}{10}\ci{+1.82}{0.13}{3.58} & $+684$ & $+4.3\%$ \\
\FT $\to$ \KP & \ci{82.42}{79.36}{85.42} & \effectshade{-0.20}{10}\ci{-0.20}{-2.02}{1.69} & \effectshade{+2.21}{10}\ci{+2.21}{0.46}{4.10} & $+464$ & $+6.1\%$ \\
\FT $\to$ \SU & \ci{81.51}{78.45}{84.51} & \effectshade{-1.11}{10}\ci{-1.11}{-2.86}{0.65} & \effectshade{+1.30}{10}\ci{+1.30}{-0.33}{2.93} & $+599$ & $+2.2\%$ \\
\rowcolor{PIShapeGroupGrey}
\multicolumn{6}{l}{\textbf{Prompt templates}\quad alternative templates for the teacher and student} \\
\FT swap & \ci{80.99}{77.80}{84.05} & \effectshade{-1.63}{10}\ci{-1.63}{-3.39}{0.13} & \effectshade{+0.78}{10}\ci{+0.78}{-1.17}{2.80} & $+568$ & $-1.5\%$ \\
\CS swap & \ci{83.92}{80.86}{86.85} & \effectshade{+0.13}{10}\ci{+0.13}{-1.50}{1.76} & \effectshade{+3.71}{10}\ci{+3.71}{2.02}{5.47} & $+1{,}068$ & $-1.2\%$ \\
\rowcolor{PIShapeGroupGrey}
\multicolumn{6}{l}{\textbf{Correction markers}\quad loss excluded or downweighted at sampled marker positions} \\
\FT exclude & \ci{81.97}{78.91}{84.83} & \effectshade{-0.65}{10}\ci{-0.65}{-2.41}{1.17} & \effectshade{+1.76}{10}\ci{+1.76}{-0.07}{3.58} & $-164$ & $+0.4\%$ \\
\FT downweight & \ci{81.84}{78.71}{84.83} & \effectshade{-0.78}{10}\ci{-0.78}{-2.34}{0.78} & \effectshade{+1.63}{10}\ci{+1.63}{-0.07}{3.32} & $-64$ & $-0.3\%$ \\
\CS exclude & \ci{83.07}{79.95}{86.00} & \effectshade{-0.72}{10}\ci{-0.72}{-2.21}{0.78} & \effectshade{+2.86}{10}\ci{+2.86}{1.04}{4.69} & $+231$ & $+0.3\%$ \\
\CS downweight & \ci{83.07}{80.08}{86.00} & \effectshade{-0.72}{10}\ci{-0.72}{-2.34}{0.91} & \effectshade{+2.86}{10}\ci{+2.86}{1.04}{4.69} & $+132$ & $+1.1\%$ \\
\midrule
\rowcolor{PIShapeGroupGrey}
\multicolumn{6}{l}{\textbf{Loss support}\quad thinking-enabled \FT, each alternative minus \textsc{Early-1K}} \\
\multicolumn{6}{@{}l@{}}{%
\begin{tabular*}{\linewidth}{@{}l@{\extracolsep{\fill}}GGG@{}}
 & \textbf{Step 25} & \textbf{Step 50} & \textbf{Step 100} \\
First-4K & \effectshade{+0.98}{10}\ci{+0.98}{-0.91}{2.86} & \effectshade{-0.85}{10}\ci{-0.85}{-2.93}{1.17} & \effectshade{-1.56}{10}\ci{-1.56}{-3.91}{0.72} \\
Distributed-1K & \effectshade{+0.98}{10}\ci{+0.98}{-0.91}{2.86} & \effectshade{+0.52}{10}\ci{+0.52}{-1.76}{2.80} & \effectshade{-1.30}{10}\ci{-1.30}{-3.45}{0.91} \\
\end{tabular*}} \\
\bottomrule
\end{tabular}
\end{table}

\begin{figure}[!t]
\centering
\includegraphics[width=\linewidth]{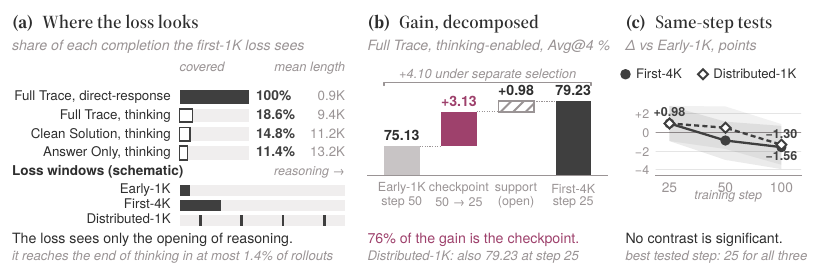}
\caption{\textbf{Most of the apparent loss-window gain comes from checkpoint choice.}
(a) Mean completion coverage of the inherited loss window and schematic alternatives.
(b) Checkpoint and loss-window contributions to the development-selected gain.
(c) Same-step contrasts with \textsc{Early-1K}, with 95\% problem-cluster bootstrap intervals.
The loss-window experiments use one training seed.}
\label{fig:audit}
\end{figure}

\noindent\textbf{Checkpoint choice explains most of the apparent loss-window gain.}
\label{sec:support-mismatch}
Beyond individual marker positions, we test whether long thinking-enabled rollouts benefit from supervision beyond their opening.
For thinking-enabled \FT training, \textsc{First-4K} extends the inherited \textsc{Early-1K} window to 4,096 tokens, while \textsc{Distributed-1K} places four 256-token windows across the reasoning (Figure~\ref{fig:audit}a).

Both alternatives exceed development-selected \textsc{Early-1K} by about four points, but stop at step 25 rather than its step 50.
At common step 25 the apparent advantage falls below one point.
The intervals for both alternatives' gains include zero at every matched checkpoint (Figure~\ref{fig:audit}b--c), leaving stopping time as the main source of the apparent improvement \citep{dodge2020finetuning,bouthillier2021variance}.

\section{Related Work}
\label{sec:related-work}

\noindent\textbf{Learning from privileged context.}
On-policy distillation trains on the student's own responses, bringing teacher feedback to the states it actually visits \citep{agarwal2024onpolicy}.
Conditioning the teacher on demonstrations \citep{shenfeld2026selfdistillation} or solutions \citep{zhao2026opsd} extends this idea to self-distillation, connecting it to learning using privileged information \citep{vapnik2009new,lopezpaz2016unifying}.
This makes additional context a source of supervision without requiring a separate, stronger model.
Its usefulness nevertheless depends on teacher--student compatibility and the training objective, now central questions in studies of on-policy distillation \citep{li2026rethinkingopd,zhu2026manyfaces}.

\noindent\textbf{What the reference contributes.}
Concurrent work is examining what privileged references contribute beyond distillation itself.
Comparisons of reference types and reasoning modes find gains without target-specific solutions \citep{shrestha2026rethinking,ichihara2026op2sd}, while structured guidance can outperform full traces \citep{zhao2026morepi}.
We construct \textsc{AMPLE-Math} to hold answer availability fixed across six reasoning views.
Using this reusable suite, we follow each reference's benefit beyond reference-free training across model families and checkpoints.
Holding the references fixed, we also show that changing student training trajectories can reverse transfer in both families.

\noindent\textbf{From teacher signals to student learning.}
Closer teacher imitation does not always improve generalization \citep{stanton2021does}, making it important to understand what the teacher's scores encourage.
Work on OPSD identifies possible obstacles in preferences for the supplied solution \citep{harne2026biased}, likelihood shifts without useful token credit \citep{nguyen2026likelihood}, and suppressed reconsideration \citep{kaur2026rethinking}.
Complementary methods revise token targets, hidden-state supervision, or the teacher itself \citep{shen2026purified,li2026phf,feng2026past,sun2026sropsd}.
We test whether changing these signals changes student behavior, pairing profiles on fixed and trained-student responses with marker-loss and loss-window interventions.
Matching checkpoints also separates the effect of loss design from the benefit of stopping earlier.

\section{Discussion and Conclusion}
\label{sec:discussion}

Successful distillation and a useful reference are different outcomes.
Our answer-matched comparisons show that much of the improvement can occur without a reference, while a reference's additional benefit depends on the model and training stage.
More complete solutions do not consistently yield greater gains.
Replacing genuine references with unrelated ones can reduce accuracy, so reference-free gains do not make reference content irrelevant.
Separating these effects makes reference selection a question of what helps the student beyond distillation itself.

The student's own responses are central to that choice because they determine the prefixes the teacher supervises.
With the teacher and references unchanged, switching from short direct-response rollouts to long thinking-enabled rollouts under the same loss window turns gains into losses in both families.
An asymmetry remains in the direct-response training configuration even without privileged information, since a thinking-enabled teacher supervises direct-response attempts.
Learning from these scores may make existing reasoning capabilities more accessible through parameters shared by both inference modes.
The concentration of gains on problems the base already sometimes solves is consistent with this interpretation.

Teacher profiles show how feedback varies across student responses and identify candidate changes to supervision.
Relaxing loss at sampled correction markers barely changes their use, while matching checkpoints attributes most of the apparent benefit of broader loss coverage to stopping earlier.
Changing the training signal does not necessarily produce the intended change in the student.
Our evidence comes from short-run LoRA training on mathematics in two model families.
These findings make reference design and student training a joint problem.
They motivate testing whether reference content that adapts to the student's evolving attempts improves on a fixed representation throughout training.
\textsc{AMPLE-Math} provides a reusable setting for studying that interaction while holding answer information fixed.
Combined with teacher profiles and matched interventions, it enables future OPSD research to test both what a reference adds and the explanations proposed for its effects.
The central lesson is to judge privileged information by what it adds to the student, not by how complete a solution it gives the teacher.

\clearpage
\subsection*{AI use statement}
The authors determined the research questions, method, experimental design, analysis, and claims.
Generative AI tools assisted with manuscript and software preparation.
Their feedback helped the authors iteratively refine hypotheses, experimental design, mathematical formulation, and interpretation of results.
Appendix~\ref{app:dataset} describes their use in generating and auditing the dataset's reasoning views.
The authors edited and checked the manuscript, verified the reported results, and take responsibility for the work.

\subsection*{Ethics statement}
This work uses publicly released mathematical reasoning data and open model checkpoints and involves no human subjects or personal data.
\textsc{AMPLE-Math} preserves source provenance, and its release is subject to the licenses of its upstream data and models.
As with other reasoning-model research, the trained models can generate incorrect or misleading outputs; our evaluation is limited to mathematical correctness and should not be interpreted as a general safety assessment.

\subsection*{Reproducibility statement}
Appendix~\ref{app:experimental-details} gives the objective, optimization, checkpoint and decoding settings, and statistical estimators.
Appendix~\ref{app:dataset} describes dataset construction and split rules, and Appendices~\ref{app:transfer-evidence} to \ref{app:registered-stress-tests} provide the complete result tables.
Links to the code on \coderepo{} and \textsc{AMPLE-Math} on \datasetrepo{} are provided on the first page.

\bibliography{iclr2027_conference}
\bibliographystyle{iclr2027_conference}

\appendix
\begingroup
\section{Common Experimental Details}
\label{app:experimental-details}

\subsection{Implementation and optimization}

Qwen3-1.7B uses a frozen backbone and LoRA adapters ($r=64$, $\alpha=128$, dropout $0$) on the query, key, value, output, gate, up, and down projections.
Training uses effective batch size 32, learning rate $5\times10^{-6}$, AdamW \citep{loshchilov2019decoupled} with $\beta_2=0.999$, rollout and loss temperatures 1.1, and 100 optimizer steps.
Checkpoints are saved at steps $0,1,2,5,10,15,20,25,50,75,100$.
Direct-response rollouts are capped at 1,024 tokens, all supervised, while thinking-enabled Qwen rollouts extend to 20,764 tokens with the same first-1,024-token loss support unless an intervention changes $I(y)$.
We follow the released code's generalized divergence at $\beta=0$ with a per-term clip of $0.05$, rather than the $\beta=0.5$ described in the paper \citep{zhao2026opsd}.

SmolLM3-3B uses the same recipe with native thinking/no-thinking templates, a 24,536-token thinking-enabled rollout cap, and a 40,960-token training context. Its loss still covers the first 1,024 completion tokens.
In both models the teacher is the frozen, thinking-enabled backbone with the student adapter disabled.
Student rollouts and evaluation receive no PI. We evaluate the same trained checkpoints with thinking enabled and under direct-response generation.

\subsection{Evaluation and effective correctness}

In-domain evaluation draws four samples per problem on the frozen 384-problem test split at temperature 1.0, top-$p=0.95$, and top-$k=20$.
Both families use a 32,768-token evaluation context and an initial 16,384-token generation budget.
Every capped response is regenerated from the beginning with the same sampling seed, whether or not it already contains an answer.
For a $P$-token prompt, the new-token limit is $\min(32{,}512,32{,}768-P-256)$, reserving a 256-token margin.
\emph{Effective correctness} grades the extended-budget output when available, otherwise the initial response.
External evaluation follows the released OPSD prompt and sampling protocol: 12 samples per problem, temperature 1.0, top-$p=0.95$, and 38,912 generated tokens within a 40,960-token context.
The external rescue trigger is inactive at this budget, so main-pass and rescue-aware scores coincide.

For saved-prefix comparisons, we decode the first 4,096, 8,192, or 16,384 token IDs from each initial response and apply the same answer grader.
Shorter responses are graded in full. Only the full-evaluation endpoint uses the extended-budget output when available.
These are grades of stored prefixes, not new generations asked to finish within a smaller budget.

Training-configuration comparisons primarily use common steps 25, 50, and 100, with development-selected checkpoints as secondary summaries.
Development-selected checkpoints maximize Avg@2 on the 192-problem development split over steps 25, 50, 75, and 100, without test access.
The selected steps for the seed-0 six-view comparison are 75, 50, 75, 75, 100, and 25 for \AO, \GI, \KP, \CS, \SU, and \FT, respectively.

\subsection{Profile statistics}

Writing the token shift as $\delta_{v,t}(a)=\log p^T_{v,t}(a)-\log p^S_t(a)$, correction pressure $F_v$ averages $\delta_{v,t}(y_t)$ over sampled marker positions within each problem, then weights contributing problems equally.
Temporal KL allocation reports the share of full-vocabulary $\KL(p^T_{v,t}\,\|\,p^S_t)$ in each quarter of the retained span.
These statistics and correctness alignment are bootstrapped over problems.

For the correction-pressure diagnostic, token pieces are lowercased after replacing tokenizer space markers with spaces, then matched by substring against \emph{wait}, \emph{but}, \emph{however}, \emph{maybe}, \emph{actually}, \emph{recheck}, \emph{check}, \emph{verify}, \emph{reconsider}, \emph{mistake}, and \emph{instead}.
This lexical rule identifies sampled marker tokens rather than all semantically corrective continuations.

\subsection{Statistical estimators}

The problem is the inferential unit.
Avg@$n$ first averages the $n$ sampled correctness indicators within problem and then averages across problems.
Paired effects are differences of per-problem averages.
We construct 95\% percentile intervals from 10,000 problem-cluster bootstrap resamples, reusing one resampled problem-index array across compared conditions.
For balanced in-domain tests, resampling is stratified by the construction strata.
Pooled external intervals resample all 90 problems together, without benchmark stratification.
We distinguish paired intervals excluding zero from effects surviving Holm correction \citep{holm1979simple} across a family of comparisons.
Intervals including zero do not establish equivalence.
Reported $p$-values are two-sided bootstrap tail probabilities, twice the smaller of the resampled fractions at or below and at or above zero, floored at $1/10{,}000$.

Seed-aware aggregates first average the paired effects within problem across the observed seeds, then bootstrap the resulting problem vector.
These intervals describe problem-sampling uncertainty for the observed seeds.
Training-seed variability is reported separately through per-seed effects and seed-level $t$ intervals.
Paired seed-level checks use the four matched \AO-minus-\FT differences, while the \CS-minus-reference-free check uses a Welch interval over three seed means per configuration.

We apply Holm's procedure to the six view-versus-base tests, the six in-domain view-versus-reference-free comparisons, and the two external comparisons of \AO and \FT with the reference-free control.
For interventions, we correct within each control family (four marker-mask, two other-problem, three trace-opening, and two prompt-template comparisons), across all eleven controls, and across the six loss-support contrasts.

\subsection{Release materials and computation}
\label{app:release}

The code and \textsc{AMPLE-Math} dataset are linked on the first page through \coderepo{} and \datasetrepo{}, respectively.
Row-level evaluation outputs are available on request.
We estimate total project compute at approximately 1,000 H100 GPU-hours, including training, evaluation, teacher profiling, dataset construction, and setup.

\section{AMPLE-Math Construction and Auditing}
\label{app:dataset}

\subsection{Source, filtering, and the frozen population}
\label{app:pi-shape-source}

The source is the \texttt{metadata} configuration of OpenThoughts-114k \citep{guha2025openthoughts}: one problem, one verified answer, and one natural reasoning trace per row.
We required a gradable answer and a source trace whose final answer agrees with it, collapsed within-corpus near duplicates, excluded source traces above 16,384 tokens, and screened for exact and semantic overlap with MATH-500 \citep{lightman2023verify}, AIME 2024, and JEEBench \citep{arora2023jeebench}.
We then selected 5,480 eligible questions for the frozen pool.
Construction lengths use the fast Llama-3.1-8B-Instruct tokenizer without special tokens.
A question enters the complete resource only if all six views exist, every generated view passes surface validation and source-fidelity review, and body lengths satisfy \AO $<$ \GI $<$ \KP $<$ \SU $<$ \FT.
\CS is off this length spine and is not constrained relative to the generated views.
In total, 5,319 do, giving $5{,}319\times6=31{,}914$ supervision records and leaving 161 excluded questions.

A word-shingle scan of the 90 external evaluation problems (AIME 2024, AIME 2025, and HMMT February 2025) against the 5,319 complete resource problems found no exact or near duplicates.
The largest 8-gram containment of any benchmark problem was $0.25$, from shared contest phrasing.
Because \textsc{AMPLE-Math} and the released OPSD training data draw on the same upstream source, 202 of the 384 test problems (52.6\%) also appear in that released data.
Only the original-data implementation check trains on those data, and it is evaluated on the external benchmarks rather than the overlapping in-domain test set.
Every representation and rollout comparison uses the local resource with disjoint train, development, and test splits.

\subsection{The six answer-matched views}
\label{app:pi-shape-views}

\AO has no reasoning body, while \CS and \FT use the source's polished solution and complete trace.
Qwen3.6-35B-A3B-FP8 compresses the intact trace into a short paragraph for \GI, ordered steps for \KP, and a narrative for \SU, without a separate verified-answer field \citep{qwen2026qwen36}.
Up to five reproducible, seeded attempts use temperatures $0.0$, $0.3$, $0.6$, $0.9$, and $1.2$.
Candidates undergo surface validation for completion, form, and prompt leakage, then source-fidelity review for unsupported claims, backward-constructed reasoning, and confidence upgrades.
Every accepted view passes both stages.
Fidelity review preserves the source's reasoning and uncertainty rather than certifying or repairing its proof.
Generation and review use the same model under different prompts, so the review may share the generator's blind spots.

All views share the canonical answer but vary the reasoning's representation, wording, and detail.
Records render as \verb|<think>| body \verb|</think><answer>| answer \verb|</answer>|, with exactly one boxed answer in the answer section.
The aligned example shows how the six views express the same solution.

\begin{table}[!htb]
\textbf{An aligned example.}\par\smallskip
\begingroup
\tablefont
\setlength{\fboxsep}{7pt}
\noindent\colorbox{PIShapeGroupGrey}{%
  \parbox{\dimexpr\linewidth-2\fboxsep\relax}{%
    \textbf{Shared problem}\quad What is the value of $6\times(5-2)+4$?\par
    \smallskip
    \textbf{Canonical answer}\quad $22$
  }%
}\par\smallskip
\setlength{\tabcolsep}{0pt}
\renewcommand{\arraystretch}{1.04}
\arrayrulecolor{PIShapeCharcoal!25}
\begin{tabularx}{\linewidth}{@{}>{\raggedright\arraybackslash}p{96pt}@{\hspace{10pt}}>{\raggedright\arraybackslash}X@{}}
\textbf{PI view} & \textbf{Reasoning body (excerpts)} \\
\midrule
\addlinespace[3pt]
\leavevmode\viewsq{PIViewAO}\AO
& \textit{Empty reasoning body.} \\
\addlinespace[5pt]
\midrule
\addlinespace[3pt]
\leavevmode\viewsq{PIViewGI}\GI
& Apply order of operations: evaluate parentheses, then multiply, then add. \\
\addlinespace[5pt]
\midrule
\addlinespace[3pt]
\leavevmode\viewsq{PIViewKP}\KP
& Key steps:\newline
  -- Apply order of operations, starting with parentheses \texttt{[...]}\newline
  -- Complete the final addition: 18 plus 4 equals 22 \\
\addlinespace[5pt]
\midrule
\addlinespace[3pt]
\leavevmode\viewsq{PIViewCS}\CS
& The value of $6\times(5-2)+4$ is calculated as follows:\newline
  1. \textbf{Parentheses first}: $5-2=3$. \texttt{[...]}\newline
  3. \textbf{Add}: $18+4=22$.\newline
  \textbf{Final Answer:} $\boxed{22}$ \\
\addlinespace[5pt]
\midrule
\addlinespace[3pt]
\leavevmode\viewsq{PIViewSU}\SU
& Summary: I begin by identifying the expression as a standard arithmetic problem requiring strict adherence to the order of operations, \texttt{[...]} leading to the final value of twenty-two. \\
\addlinespace[5pt]
\midrule
\addlinespace[3pt]
\leavevmode\viewsq{PIViewFT}\FT
& Okay, let's see. I need to figure out the value of 6 times (5 minus 2) plus 4.
  Hmm, order of operations, right? I remember learning PEMDAS--Parentheses, Exponents, Multiplication and Division, Addition and Subtraction.
  \texttt{[...]} Wait, let me double-check to make sure I didn't skip any steps.
  \texttt{[...]} Yep, 22 should be the correct answer. \\
\addlinespace[3pt]
\bottomrule
\end{tabularx}
\arrayrulecolor{black}
\par\smallskip
Original wording is preserved, with Markdown and math typeset. Omissions are marked \texttt{[...]}.
\endgroup
\end{table}

\subsection{Qwen and SmolLM3 split construction}
\label{app:pi-shape-splits}

We drew the train, development, and test sets from 3,399 candidates remaining after 1,920 prior exclusions from the 5,319 complete problems, keeping the profiling set disjoint from all three splits.
For Qwen3-1.7B, four problem-only direct-response samples from the frozen base define high-success ($p\ge0.75$), intermediate ($p\in\{0.25,0.5\}$), and zero-success ($p=0$) bands.
A zero-success problem is eligible for the third band only when at least one structured PI teacher generation is correct.
The final split contains 512/64/128 problems per band in train/dev/test, for totals $1{,}536/192/384$.

SmolLM3 uses the same development and test IDs but a training split rebuilt from its own direct-response outcomes.
The same structured-PI recoverability rule permits 512 training problems per band.
On the shared test IDs, SmolLM3 has 129 high-success, 97 intermediate, and 158 zero-success problems.
Overall performance on all 384 is primary, with these unequal bands used for secondary analysis.

\subsection{Difficulty and view lengths}
\label{app:pi-shape-difficulty}

Dataset difficulty is distinct from the student-relative outcomes used for the experimental splits.
Three evaluator models (Qwen2.5-1.5B-Instruct, Llama-3.1-8B-Instruct, and Qwen3.6-35B-A3B-FP8, \citealp{qwen2024qwen25,grattafiori2024llama3,qwen2026qwen36}) answered each of the 5,480 frozen questions from the statement alone, with 8 to 16 attempts each.
A binomial Rasch model \citep{rasch1960probabilistic} jointly fits evaluator abilities and question difficulties to each evaluator's correct and actually graded counts, rather than the nominal number of attempts.
A zero-centered $L_2$ penalty of $0.5$ keeps all-correct and all-incorrect items finite.
Difficulty measures final-answer success under a fixed decoding and grading protocol, not proof quality.
Figure~\ref{fig:pi-shape-difficulty} shows the difficulty distribution and its relation to source-trace length.
Figure~\ref{fig:pi-shape-view-lengths} compares reasoning-body lengths across views.

\begin{figure}[!htb]
\centering
\begin{minipage}[t]{0.47\linewidth}
\centering
\includegraphics[width=\linewidth]{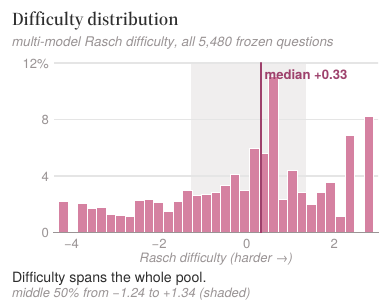}
\end{minipage}\hfill
\begin{minipage}[t]{0.47\linewidth}
\centering
\includegraphics[width=\linewidth]{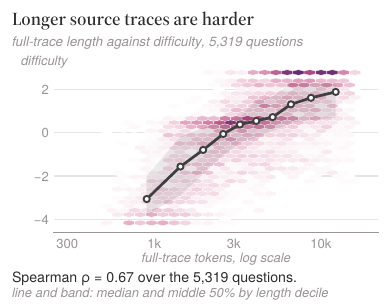}
\end{minipage}
\caption{\textbf{Difficulty and source-trace length in \textsc{AMPLE-Math}.}
Left: multi-model Rasch difficulty over the 5,480-question frozen pool.
Right: full-trace token length versus difficulty for the 5,319 complete six-view problems.
Hexagon shading counts problems, and the line and band show median difficulty and its interquartile range within source-trace length deciles.}
\label{fig:pi-shape-difficulty}
\end{figure}

\begin{figure}[!htb]
\centering
\includegraphics[width=\linewidth]{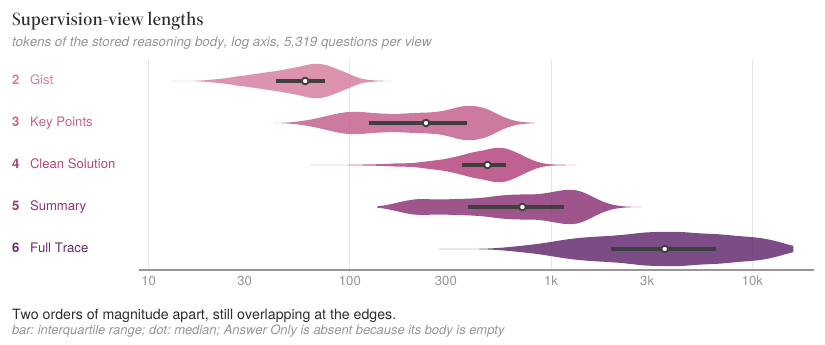}
\caption{\textbf{Reasoning-body lengths across the five nonempty views.}
Token-length distributions of the stored reasoning bodies over the 5,319 complete questions, on a logarithmic axis, with the interquartile range and median marked; \AO is absent because its body is empty by construction.
These lengths use the Llama-3.1-8B-Instruct construction tokenizer. Main-text lengths measure the rendered privileged reference under the Qwen3 tokenizer.}
\label{fig:pi-shape-view-lengths}
\end{figure}

\begin{samepage}
\section{Student-Transfer Evidence}
\label{app:transfer-evidence}

The evidence follows the main paper's distinction between distillation gains and reference contributions, then checks where the gains occur and how they depend on training.
Secondary direct-response and saved-prefix analyses appear at the end.
All evaluation is unprivileged and follows the common protocol in Appendix~\ref{app:experimental-details}.
\par
\end{samepage}

\subsection{Reference contribution in-domain and externally}
\label{app:teacher-controls}

The reference-free teacher is the frozen thinking-enabled backbone scoring the student's prefix without PI.
The wrong-answer control uses \AO with the canonical answer replaced by another problem's answer in the same format. No training problem retains its own answer.
Both controls use three training seeds.
At step 100, the three-seed \CS comparison gives $+1.30\,[0.20,2.41]$ points over the three-seed reference-free control.
This small unadjusted advantage does not survive Holm correction across the six-view family.

Tables~\ref{tab:teacher-controls} and~\ref{tab:view-minus-control} give the Qwen controls and all six views' contrasts against each control.
In this six-view comparison, view rows use seed 0 while each control averages its three seeds.
Only \CS's unadjusted interval over the reference-free student lies above zero (raw $p=0.024$, Holm-adjusted $0.15$).

\begin{table}[!htb]
\caption{\textbf{Teacher controls.}
$\Delta$ Avg@4 versus the frozen base with paired 95\% intervals at common checkpoints.
Three-seed rows average per-problem effects over seeds 0--2.}
\label{tab:teacher-controls}
\centering
\tablesetup
\setlength{\tabcolsep}{5pt}
\begin{tabular}{@{}lrrr@{}}
\toprule
\textbf{Control} & \textbf{Step 25} & \textbf{Step 50} & \textbf{Step 100} \\
\midrule
Reference-free (seed 0) & \ci{+1.63}{0.00}{3.26} & \ci{+3.26}{1.50}{5.08} & \ci{+1.76}{-0.20}{3.71} \\
Reference-free (3 seeds) & -- & \ci{+2.89}{1.39}{4.45} & \ci{+1.80}{0.20}{3.47} \\
\AO, wrong answer (seed 0) & \ci{+1.30}{-0.52}{3.12} & \ci{+2.28}{0.59}{3.97} & \ci{+0.98}{-1.04}{2.99} \\
\AO, wrong answer (3 seeds) & -- & \ci{+3.06}{1.58}{4.60} & \ci{+1.67}{0.07}{3.32} \\
\bottomrule
\end{tabular}
\end{table}

\begin{table}[!htb]
\caption{\textbf{Views against the controls.}
Each seed-0 view minus each three-seed control at step 100, in Avg@4 points with unadjusted paired 95\% intervals.}
\label{tab:view-minus-control}
\centering
\tablesetup
\setlength{\tabcolsep}{5pt}
\begin{tabular}{@{}lrr@{}}
\toprule
\textbf{View} & \textbf{minus reference-free} & \textbf{minus wrong-answer \AO} \\
\midrule
\viewsq{PIViewAO}\AO & \ci{+0.35}{-1.04}{1.71} & \ci{+0.48}{-0.85}{1.76} \\
\viewsq{PIViewGI}\GI & \ci{+0.35}{-1.13}{1.80} & \ci{+0.48}{-0.93}{1.87} \\
\viewsq{PIViewKP}\KP & \ci{+1.00}{-0.48}{2.45} & \ci{+1.13}{-0.37}{2.58} \\
\viewsq{PIViewCS}\CS & \ci{+1.78}{0.24}{3.30} & \ci{+1.91}{0.52}{3.30} \\
\viewsq{PIViewSU}\SU & \ci{+0.74}{-0.69}{2.17} & \ci{+0.87}{-0.52}{2.24} \\
\viewsq{PIViewFT}\FT & \ci{+0.61}{-0.87}{2.11} & \ci{+0.74}{-0.67}{2.15} \\
\bottomrule
\end{tabular}
\end{table}

Table~\ref{tab:external-sixview} gives the external counterpart to main Table~\ref{tab:profile-prediction}, comparing each PI-trained student with the reference-free student rather than the frozen base.
No external view-minus-control interval lies entirely above zero.
The unadjusted \FT-minus-control interval lies below zero, but the contrast does not survive Holm correction over the two extreme-view tests.

\begin{table}[!htb]
\caption{\textbf{External view-minus-control comparisons.}
Step-50 Avg@12 differences against the reference-free student, with unadjusted paired 95\% intervals over 90 problems.
\AO and \FT average four training seeds, the control three, and the other views use one.
Absolute effects and benchmark breakdowns are in Table~\ref{tab:profile-prediction}.}
\label{tab:external-sixview}
\centering
\tablesetup
\begin{tabular}{@{}lr@{}}
\toprule
\textbf{View} & \textbf{Difference vs.\ reference-free student} \\
\midrule
\viewsq{PIViewAO}\AO & \ci{-0.35}{-1.64}{1.00} \\
\viewsq{PIViewGI}\GI & \ci{+0.28}{-1.94}{2.47} \\
\viewsq{PIViewKP}\KP & \ci{-0.46}{-2.96}{1.98} \\
\viewsq{PIViewCS}\CS & \ci{+0.83}{-1.05}{2.78} \\
\viewsq{PIViewSU}\SU & \ci{-0.46}{-2.53}{1.54} \\
\viewsq{PIViewFT}\FT & \ci{-1.78}{-3.43}{-0.21} \\
\bottomrule
\end{tabular}
\end{table}

The SmolLM3 comparison keeps steps 50 and 100 together with per-seed gains and completion diagnostics (Table~\ref{tab:smollm3-reference-modes}).
\FT improves on the reference-free students with thinking enabled at both checkpoints, but by step 100 all three configurations fall below the frozen base.
The table also retains direct-response outcomes, where the late deterioration is especially pronounced.
Even after re-generation, the reference-free and \AO students' direct-response no-answer rates rise from $20.2\%$ and $22.0\%$ at step 50 to $42.0\%$ and $39.4\%$ at step 100.
Table~\ref{tab:smollm3-reference-modes} separates final-output accuracy from initial-pass completion diagnostics. For thinking-enabled \FT at step 50, re-generation reduces the no-answer rate from $19.7\%$ to $1.3\%$.

\begin{table}[!htb]
\caption{\textbf{SmolLM3-3B reference comparisons across checkpoints and evaluation modes.}
All trained configurations use direct-response training and three seeds.
\textbf{A:} Final-output Avg@4 gains over the frozen base, in percentage points with 95\% problem-cluster bootstrap intervals. Parenthesized triplets give gains for seeds 0, 1, and 2.
\textbf{B:} Paired \FT-minus-reference-free contrasts with the same interval convention.
\textbf{C:} Initial-pass mean lengths, 16,384-token cap rates, and no-answer rates, before re-generation. No answer means no extractable \texttt{\textbackslash boxed\{...\}} answer. Extraction uses the last boxed answer in the text. Diagnostics pool 4,608 responses per trained configuration.
Frozen-base Avg@4 is $82.16\%$ with thinking enabled and $40.23\%$ under direct-response evaluation.}
\label{tab:smollm3-reference-modes}
\centering
\tablesetup
\newcommand{\smolgain}[4]{\begin{tabular}[t]{@{}r@{}}\ci{#1}{#2}{#3}\\(#4)\end{tabular}}
\begin{tabular*}{\linewidth}{@{\extracolsep{\fill}}lrr@{}}
\toprule
\multicolumn{3}{@{}l}{\color{PIShapeRoseDark}\textbf{A. Accuracy gains versus the frozen base}} \\
\textbf{Student} & \textbf{Thinking-enabled} & \textbf{Direct-response} \\
\midrule
\multicolumn{3}{@{}l}{\textbf{Step 50}} \\
Reference-free & \smolgain{+0.59}{-1.06}{2.26}{$+0.20, +0.46, +1.11$} & \smolgain{+13.24}{9.96}{16.49}{$+12.76, +11.65, +15.30$} \\
\viewsq{PIViewAO}\AO & \smolgain{+0.87}{-0.74}{2.50}{$+0.46, +1.37, +0.78$} & \smolgain{+11.91}{8.57}{15.34}{$+12.11, +12.50, +11.13$} \\
\viewsq{PIViewFT}\FT & \smolgain{+2.58}{1.06}{4.14}{$+2.93, +2.41, +2.41$} & \smolgain{+2.60}{0.46}{4.77}{$+1.56, +3.58, +2.67$} \\
\addlinespace[3pt]
\multicolumn{3}{@{}l}{\textbf{Step 100}} \\
Reference-free & \smolgain{-5.25}{-7.20}{-3.32}{$-4.75, -6.25, -4.75$} & \smolgain{-10.39}{-13.15}{-7.68}{$-11.72, -9.70, -9.77$} \\
\viewsq{PIViewAO}\AO & \smolgain{-5.03}{-6.94}{-3.17}{$-5.14, -4.88, -5.08$} & \smolgain{-10.09}{-12.85}{-7.38}{$-11.78, -12.76, -5.73$} \\
\viewsq{PIViewFT}\FT & \smolgain{-2.19}{-4.08}{-0.37}{$-3.97, -1.24, -1.37$} & \smolgain{+2.76}{0.39}{5.14}{$-0.72, +5.21, +3.78$} \\
\midrule
\multicolumn{3}{@{}l}{\color{PIShapeRoseDark}\textbf{B. Full Trace minus reference-free}} \\
Step 50 & \ci{+2.00}{0.82}{3.17} & \ci{-10.63}{-13.19}{-8.07} \\
Step 100 & \ci{+3.06}{1.71}{4.41} & \ci{+13.15}{10.94}{15.39} \\
\bottomrule
\end{tabular*}

\vspace{5pt}
\setlength{\tabcolsep}{2.5pt}
\begin{tabular*}{\linewidth}{@{\extracolsep{\fill}}lrrrrrrr@{}}
\toprule
\multicolumn{8}{@{}l}{\color{PIShapeRoseDark}\textbf{C. Initial-pass length and completion diagnostics}} \\
& & \multicolumn{3}{c}{\textbf{Thinking-enabled}} & \multicolumn{3}{c}{\textbf{Direct-response}} \\
\cmidrule(lr){3-5}\cmidrule(lr){6-8}
\textbf{Student} & \textbf{Step} & \textbf{Tokens} & \textbf{Cap} & \textbf{No answer} & \textbf{Tokens} & \textbf{Cap} & \textbf{No answer} \\
\midrule
Frozen base & 0 & 7,492 & $12.9\%$ & $10.9\%$ & 833 & $0.3\%$ & $0.8\%$ \\
\midrule
Reference-free & 50 & 9,078 & $19.1\%$ & $18.3\%$ & 5,760 & $2.9\%$ & $21.3\%$ \\
\viewsq{PIViewAO}\AO & 50 & 9,485 & $20.5\%$ & $19.0\%$ & 5,959 & $3.0\%$ & $23.4\%$ \\
\viewsq{PIViewFT}\FT & 50 & 9,746 & $22.6\%$ & $19.7\%$ & 1,809 & $0.6\%$ & $3.3\%$ \\
\addlinespace[3pt]
Reference-free & 100 & 8,700 & $15.1\%$ & $17.1\%$ & 11,153 & $42.8\%$ & $52.5\%$ \\
\viewsq{PIViewAO}\AO & 100 & 8,866 & $16.2\%$ & $17.1\%$ & 10,875 & $43.2\%$ & $49.8\%$ \\
\viewsq{PIViewFT}\FT & 100 & 9,841 & $23.4\%$ & $20.8\%$ & 6,629 & $25.5\%$ & $23.5\%$ \\
\bottomrule
\end{tabular*}
\end{table}

\begin{samepage}
\subsection{Where the gains occur}

Table~\ref{tab:construction-strata} reports the step-100 multi-seed Qwen gains within the three frozen split-construction groups, matching Figure~\ref{fig:sixview}c.
The largest gains occur in the zero-correct direct-response group, where the frozen base nevertheless answers $61.91\%$ of samples correctly with thinking enabled.
Within this group, \CS exceeds reference-free training by $3.06\,[0.46,5.79]$ points, its largest additional gain among the three groups.
The intervals describe unadjusted within-group comparisons.
\par
\end{samepage}

\begin{table}[!htb]
\caption{\textbf{Multi-seed gains by split-construction group.}
Qwen3-1.7B Avg@4 changes in percentage points at step 100 after direct-response training, with thinking-enabled evaluation.
Groups contain 128 test problems each and are defined by correct answers among four frozen-base direct-response samples.
Estimates average three training seeds for reference-free and \CS, and four for \AO and \FT; subscripts give unadjusted 95\% paired problem-bootstrap intervals (10,000 resamples within each group).
The final row gives frozen-base thinking-enabled accuracy, not a gain.}
\label{tab:construction-strata}
\centering
\tablesetup
\begin{tabular*}{\linewidth}{@{\extracolsep{\fill}}lrrr@{}}
\toprule
\textbf{Configuration} & \hdrr{Zero correct}{0/4} & \hdrr{Intermediate}{1--2/4} & \hdrr{High success}{3--4/4} \\
\midrule
\multicolumn{4}{@{}l}{\textit{Gain over frozen base}} \\
Reference-free & \ci{+3.97}{0.39}{7.62} & \ci{+0.98}{-1.82}{3.91} & \ci{+0.46}{-1.17}{2.28} \\
\viewsq{PIViewAO}\AO & \ci{+4.69}{1.32}{8.25} & \ci{+2.34}{-0.24}{4.98} & \ci{+0.20}{-1.76}{2.25} \\
\viewsq{PIViewCS}\CS & \ci{+7.03}{3.52}{10.74} & \ci{+1.89}{-0.72}{4.62} & \ci{+0.39}{-1.43}{2.34} \\
\viewsq{PIViewFT}\FT & \ci{+5.66}{2.59}{8.84} & \ci{+1.51}{-1.12}{4.25} & \ci{-0.05}{-1.76}{1.81} \\
\midrule
\multicolumn{4}{@{}l}{\textit{Paired gain over reference-free}} \\
\viewsq{PIViewAO}\AO & \ci{+0.72}{-1.51}{2.99} & \ci{+1.37}{-0.26}{3.03} & \ci{-0.26}{-1.20}{0.62} \\
\viewsq{PIViewCS}\CS & \ci{+3.06}{0.46}{5.79} & \ci{+0.91}{-0.85}{2.60} & \ci{-0.07}{-0.98}{0.78} \\
\viewsq{PIViewFT}\FT & \ci{+1.69}{-0.62}{4.00} & \ci{+0.54}{-1.46}{2.54} & \ci{-0.50}{-1.33}{0.31} \\
\midrule
Frozen-base accuracy & $61.91\%$ & $83.98\%$ & $94.73\%$ \\
\bottomrule
\end{tabular*}
\end{table}

\noindent\textbf{Validation on the original training data.}
The implementation check retains the original OPSD training data and settings \citep{zhao2026opsd}, with direct-response student rollouts capped at 1,024 tokens and a thinking-enabled teacher.
Unlike the preceding \textsc{AMPLE-Math} comparisons, this check evaluates a student trained on the original data.
Table~\ref{tab:positive-control} reports step-50 Avg@12 on the external benchmarks.
Pooled Majority@12 rises by $7.78$ points while Pass@12 falls by $7.78$ points.

\begin{table}[htbp]
\caption{\textbf{Validation on the original OPSD training data.}
Thinking-enabled external evaluation at step 50. Values are Avg@12 percentages, with 95\% intervals for the changes from the frozen base.}
\label{tab:positive-control}
\centering
\tablesetup
\begin{tabular}{@{}lrrrr@{}}
\toprule
 & \textbf{AIME24} & \textbf{AIME25} & \textbf{HMMT25} & \textbf{Pooled} \\
\midrule
Frozen base & 47.22 & 39.17 & 23.89 & 36.76 \\
OPSD step 50 & 54.72 & 42.78 & 29.44 & 42.31 \\
Difference & $+7.50$ & $+3.61$ & $+5.56$ & $+5.56$ \\
95\% CI & [2.22, 12.78] & [$-2.50$, 10.56] & [$-0.56$, 12.50] & [2.22, 9.17] \\
\bottomrule
\end{tabular}
\end{table}

Grouping the 90 external problems by the frozen base's 12 thinking-enabled samples gives 27 never-solved, 49 intermittently solved, and 14 always-solved problems.
Their Avg@12 changes are $+0.62$, $+10.03$, and $-0.60$ points, respectively.
This descriptive breakdown locates the gain among intermittently solved problems. These groups differ from the direct-response construction strata in \textsc{AMPLE-Math}.

\subsection{Training-seed and checkpoint checks}
\label{app:seed-robustness}

The Qwen extreme views \AO and \FT each use four seeds under the same data, optimizer, and schedule.
Table~\ref{tab:seed-robustness} reports their paired effects and between-seed ranges, using step 100 in domain and step 50 externally.

\begin{table}[!htb]
\caption{\textbf{Training-seed robustness of the extreme-view contrast.}
Changes versus the frozen base, with paired 95\% problem-bootstrap intervals.
In-domain evaluation uses step-100 Avg@4 on 384 problems. External evaluation uses step-50 Avg@12 on 90 problems.
Seed-aware estimates average per-problem effects over four seeds before resampling.}
\label{tab:seed-robustness}
\label{tab:external-seeds}
\centering
\tablesetup
\begin{tabular}{@{}lrrrr@{}}
\toprule
& \multicolumn{2}{c}{\textbf{In domain}} & \multicolumn{2}{c}{\textbf{External}} \\
\cmidrule(lr){2-3}\cmidrule(lr){4-5}
& \textbf{\viewsq{PIViewAO}\AO} & \textbf{\viewsq{PIViewFT}\FT} & \textbf{\viewsq{PIViewAO}\AO} & \textbf{\viewsq{PIViewFT}\FT} \\
\midrule
Seed 0 & \ci{+2.15}{0.26}{4.04} & \ci{+2.41}{0.65}{4.17} & \ci{+3.70}{0.93}{6.67} & \ci{+2.87}{0.00}{5.93} \\
Seed 1 & \ci{+1.89}{0.07}{3.78} & \ci{+3.06}{1.30}{4.88} & \ci{+3.70}{0.74}{6.76} & \ci{+1.39}{-1.39}{4.26} \\
Seed 2 & \ci{+3.06}{1.24}{4.95} & \ci{+2.80}{0.98}{4.62} & \ci{+3.33}{0.37}{6.57} & \ci{+1.57}{-1.57}{4.72} \\
Seed 3 & \ci{+2.54}{0.72}{4.30} & \ci{+1.24}{-0.65}{3.06} & \ci{+4.17}{1.30}{7.22} & \ci{+3.33}{0.56}{6.11} \\
\midrule
Seed-aware mean & \ci{+2.41}{0.86}{3.99} & \ci{+2.38}{0.91}{3.86} & \ci{+3.73}{1.27}{6.41} & \ci{+2.29}{-0.16}{4.88} \\
Between-seed range & $1.17$ & $1.82$ & $0.83$ & $1.94$ \\
\midrule
\AO minus \FT & \multicolumn{2}{c}{\ci{+0.03}{-0.91}{0.98}} & \multicolumn{2}{c@{}}{\ci{+1.44}{-0.23}{3.17}} \\
\bottomrule
\end{tabular}
\end{table}

Table~\ref{tab:seed-level-checks} compares problem-bootstrap and seed-level uncertainty for the principal step-100 contrasts.
Both checks support a small unadjusted \CS advantage over reference-free training and an evaluation-mode interaction for \AO versus \FT.
The seed-level check conditions on the test set, while the problem bootstrap conditions on the observed seeds.

\begin{table}[htbp]
\caption{\textbf{Problem-level and training-seed uncertainty give consistent readings.}
Qwen3-1.7B at step 100, with Avg@4 contrasts in percentage points.
Problem intervals average observed seeds before bootstrapping problems.
Seed-level intervals use four paired differences ($t$, 3 degrees of freedom), except \CS versus reference-free, which uses three seeds each (Welch, approximately 2.3 degrees of freedom).
All intervals are 95\% and unadjusted.}
\label{tab:seed-level-checks}
\centering
\tablesetup
\setlength{\tabcolsep}{4pt}
\begin{tabular}{@{}llrr@{}}
\toprule
\textbf{Contrast} & \textbf{Evaluation} & \textbf{Problem bootstrap} & \textbf{Seed-level $t$} \\
\midrule
\CS $-$ reference-free & Thinking-enabled & \ci{+1.30}{0.20}{2.41} & \ci{+1.30}{0.27}{2.33} \\
\AO $-$ \FT & Thinking-enabled & \ci{+0.03}{-0.91}{0.98} & \ci{+0.03}{-1.61}{1.68} \\
\AO $-$ \FT & Direct-response & \ci{+5.32}{3.47}{7.16} & \ci{+5.32}{1.9}{8.7} \\
\AO $-$ \FT & Mode interaction & \ci{+5.29}{3.26}{7.32} & \ci{+5.29}{2.62}{7.96} \\
\bottomrule
\end{tabular}
\end{table}

\begin{samepage}
At step 100, the three-seed \CS contrasts against the four-seed \AO and \FT averages are $0.69\,[-0.22,1.65]$ and $0.73\,[-0.17,1.65]$ points, respectively.
At step 50, the two available \CS seeds (0 and 2) give a $1.05\,[0.07,2.06]$-point advantage over the three-seed reference-free control.
Under development selection, the \AO-minus-\FT contrast over seeds 0--2 is $-0.24\,[-1.28,0.76]$.
\par
\end{samepage}

\subsection{Training-trajectory comparisons and completion checks}

Table~\ref{tab:qwen-common-levels} gives Qwen accuracy levels and paired training-configuration gaps at common checkpoints.
All nine gaps are negative, and only \CS at step 25 has an interval that includes zero.

\begin{table}[!htb]
\caption{\textbf{Fixed-checkpoint training-trajectory comparison.}
Qwen Avg@4 under unprivileged, thinking-enabled evaluation.
DR and TH denote direct-response and thinking-enabled training rollouts. Gaps are TH minus DR, in points with paired 95\% problem-bootstrap intervals. The frozen base scores $80.21\%$.}
\label{tab:qwen-common-levels}
\label{tab:qwen-common-gaps}
\centering
\tablesetup
\begin{tabular}{@{}llrrr@{}}
\toprule
\textbf{View} & \textbf{Step} & \textbf{DR} & \textbf{TH} & \textbf{Gap [95\% CI]} \\
\midrule
\viewsq{PIViewAO}\AO & 25 & 82.55 & 79.17 & \ci{-3.39}{-5.40}{-1.43} \\
& 50 & 83.59 & 75.20 & \ci{-8.40}{-11.00}{-5.92} \\
& 100 & 82.36 & 77.28 & \ci{-5.08}{-7.23}{-2.99} \\
\addlinespace[2pt]
\viewsq{PIViewCS}\CS & 25 & 82.03 & 81.05 & \ci{-0.98}{-2.80}{0.85} \\
& 50 & 84.64 & 77.54 & \ci{-7.10}{-9.38}{-4.95} \\
& 100 & 83.79 & 77.28 & \ci{-6.51}{-8.79}{-4.23} \\
\addlinespace[2pt]
\viewsq{PIViewFT}\FT & 25 & 82.55 & 78.26 & \ci{-4.30}{-6.25}{-2.41} \\
& 50 & 82.42 & 75.13 & \ci{-7.29}{-9.44}{-5.14} \\
& 100 & 82.62 & 72.79 & \ci{-9.83}{-11.98}{-7.68} \\
\midrule
Pooled & 25 & -- & -- & \ci{-2.89}{-4.08}{-1.71} \\
& 50 & 83.55 & 75.95 & \ci{-7.60}{-9.31}{-5.95} \\
& 100 & 82.92 & 75.78 & \ci{-7.14}{-8.66}{-5.64} \\
\bottomrule
\end{tabular}
\end{table}

\begin{samepage}
The \FT comparison is repeated on SmolLM3 with two matched training seeds (Table~\ref{tab:smollm3-seeds}).
At step 50, mean direct-response-trained and thinking-enabled-trained accuracies are $84.83\%$ and $77.99\%$, against a base of $82.16\%$.
\FT's final-output cap rate is at most $3.9\%$ in both seeds.
\par
\end{samepage}

\begin{table}[!htb]
\caption{\textbf{SmolLM3 Full Trace replication.}
Thinking-enabled-trained minus direct-response-trained Avg@4 for two matched training seeds, in points with paired 95\% intervals.}
\label{tab:smollm3-seeds}
\centering
\tablesetup
\begin{tabular}{@{}lrrr@{}}
\toprule
 & \textbf{Step 25} & \textbf{Step 50} & \textbf{Step 100} \\
\midrule
Seed 0 & \ci{-3.78}{-5.92}{-1.56} & \ci{-6.84}{-8.98}{-4.69} & \ci{-6.64}{-8.98}{-4.23} \\
Seed 1 & \ci{-3.12}{-5.01}{-1.24} & \ci{-6.84}{-8.85}{-4.82} & \ci{-9.83}{-12.11}{-7.55} \\
Seed-aware mean & \ci{-3.45}{-4.98}{-1.89} & \ci{-6.84}{-8.50}{-5.18} & \ci{-8.24}{-10.16}{-6.28} \\
Raw seed spread & $0.65$ & $0.00$ & $3.19$ \\
\bottomrule
\end{tabular}
\end{table}

\begin{samepage}
The Qwen completion checks likewise distinguish \AO from the denser views.
At development-selected checkpoints, thinking-enabled \AO training produces longer final outputs and more capping than direct-response training, yet its accuracy penalty is smaller than \FT's.
The \CS and \FT thinking-enabled students instead produce shorter responses while also losing accuracy.
The training-configuration gap thus persists after the extended-budget pass and for \FT, where capping is modest.
\par
\end{samepage}

\begin{samepage}
Table~\ref{tab:external-step50} reports the step-50 \FT training-configuration contrast by external benchmark.
The penalty's interval lies below zero on AIME 2024 and AIME 2025, but includes zero on HMMT 2025.
Against the base, the pooled effects are $-6.57\,[-9.81,-3.43]$ points for thinking-enabled training and $+2.87\,[0.00,5.93]$ for direct-response training, whose interval touches zero.
\par
\end{samepage}

\begin{table}[!htb]
\caption{\textbf{External Full Trace pair.}
Qwen Full Trace at the common step 50, with Avg@12 percentages and $\Delta$ denoting thinking-enabled-trained minus direct-response-trained.}
\label{tab:external-step50}
\centering
\tablesetup
\begin{tabular}{@{}lrrrr@{}}
\toprule
 & \textbf{AIME24} & \textbf{AIME25} & \textbf{HMMT25} & \textbf{Pooled} \\
\midrule
Frozen base & 47.22 & 39.17 & 23.89 & 36.76 \\
Direct-response trained & 53.89 & 40.56 & 24.44 & 39.63 \\
Thinking-enabled trained & 40.83 & 27.78 & 21.94 & 30.19 \\
$\Delta$ & $-13.06$ & $-12.78$ & $-2.50$ & $-9.44$ \\
95\% CI & [$-20.56$, $-5.83$] & [$-19.72$, $-6.39$] & [$-6.39$, 1.11] & [$-13.24$, $-5.83$] \\
\bottomrule
\end{tabular}
\end{table}

The thinking-enabled-trained student reaches the 38,912-token cap in 79 of 1,080 external samples, versus two for the direct-response-trained student.
Capped samples are scored as generated.
Excluding them leaves a gap of $-8.16\,[-12.16,-4.25]$ points.
Scoring every capped sample as correct, although none was, gives $-2.13\,[-6.85,+2.69]$, whose interval includes zero.

\subsection{Evaluation-mode and saved-prefix sensitivity}
\label{app:dr-mode-eval}
\label{app:output-budget}

Figure~\ref{fig:secondary-evaluation} evaluates the same checkpoints with thinking enabled and disabled, alongside the step-50 saved-prefix comparison.
Table~\ref{tab:mode-interaction-seeds} gives the four-seed Qwen \AO-minus-\FT contrast in each mode and its paired interaction at steps 50 and 100.
For SmolLM3 at step 50, the corresponding interaction is $11.02\,[8.14,13.95]$ points (Table~\ref{tab:smollm3-reference-modes}), defined as the direct-response contrast minus the thinking-enabled contrast.

\begin{figure}[!htb]
\centering
\includegraphics[width=\linewidth]{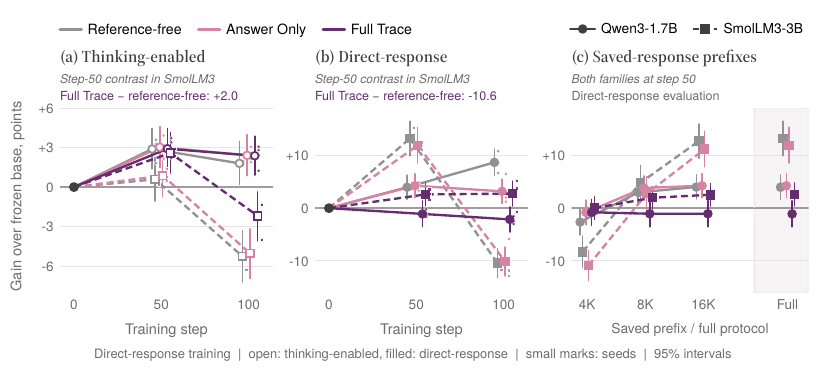}
\caption{\textbf{Student gains under both evaluation modes and saved-prefix grading.}
(a--b) Gains over the frozen base at matched checkpoints, evaluated with thinking enabled (a) and directly (b).
(c) Step-50 direct-response prefixes from the initial generation pass. Full evaluation uses the final graded outputs.
Means and 95\% problem-cluster intervals use 384 problems and three seeds, except four for Qwen \AO/\FT. Step 0 is the shared frozen base.}
\label{fig:secondary-evaluation}
\end{figure}

\begin{table}[htbp]
\caption{\textbf{Evaluation-mode interaction by training seed.}
\AO minus \FT in Avg@4 percentage points under each evaluation mode, with interaction defined as the direct-response minus thinking-enabled contrast.
Intervals are paired 95\% problem-bootstrap intervals.
Individual seed rows are at step 100. Seed-aware rows average per-problem effects over four seeds before resampling.}
\label{tab:mode-interaction-seeds}
\centering
\tablesetup
\begin{tabular}{@{}lrrr@{}}
\toprule
 & \textbf{Direct-response} & \textbf{Thinking-enabled} & \textbf{Interaction} \\
\midrule
seed 0 & \ci{+3.71}{0.98}{6.45} & \ci{-0.26}{-1.95}{1.43} & \ci{+3.97}{0.65}{7.23} \\
seed 1 & \ci{+3.52}{0.85}{6.25} & \ci{-1.17}{-2.74}{0.39} & \ci{+4.69}{1.56}{7.75} \\
seed 2 & \ci{+8.01}{5.01}{11.00} & \ci{+0.26}{-1.56}{2.08} & \ci{+7.75}{4.23}{11.13} \\
seed 3 & \ci{+6.05}{3.19}{8.98} & \ci{+1.30}{-0.20}{2.86} & \ci{+4.75}{1.56}{8.01} \\
\midrule
Step 100, seed-aware & \ci{+5.32}{3.47}{7.16} & \ci{+0.03}{-0.91}{0.98} & \ci{+5.29}{3.26}{7.32} \\
Step 50, seed-aware & \ci{+5.34}{3.69}{7.00} & \ci{+0.11}{-0.76}{0.99} & \ci{+5.22}{3.42}{7.05} \\
\bottomrule
\end{tabular}
\end{table}

\begin{samepage}
Table~\ref{tab:prefix-budget-contrasts} gives the corresponding step-100 comparison when stored Qwen direct responses are graded at progressively longer prefixes.
Both \AO and reference-free students fall below \FT at 4K, then exceed it at 8K and beyond.
The reference-free gain over the base has an interval above zero only at 16K and full evaluation, while \FT's interval includes zero at every cutoff.
\par
\end{samepage}
\begin{table}[!htb]
\caption{\textbf{Direct-response contrasts change when stored responses are graded at shorter lengths.}
Qwen3-1.7B at step 100, averaging four \AO and \FT training seeds and three reference-free seeds.
Entries are Avg@4 differences in points with paired 95\% intervals from 10,000 shared problem-cluster bootstrap resamples.
The first three rows grade decoded prefixes of 4,096, 8,192, and 16,384 main-pass tokens, without the second pass.
The final row uses the full evaluation protocol.}
\label{tab:prefix-budget-contrasts}
\centering
\tablesetup
\setlength{\tabcolsep}{2.5pt}
\begin{tabular*}{\linewidth}{@{\extracolsep{\fill}}lrrrr@{}}
\toprule
\textbf{Prefix} & \shortstack{\AO\\$-$ \FT} & \shortstack{Reference-free\\$-$ \FT} & \shortstack{Reference-free\\$-$ base} & \shortstack{\FT\\$-$ base} \\
\midrule
4K & \ci{-3.01}{-4.85}{-1.22} & \ci{-3.99}{-6.26}{-1.78} & \ci{-5.95}{-8.53}{-3.39} & \ci{-1.95}{-4.33}{0.36} \\
8K & \ci{+3.29}{1.46}{5.06} & \ci{+3.86}{1.75}{5.98} & \ci{+1.78}{-0.74}{4.25} & \ci{-2.08}{-4.46}{0.26} \\
16K & \ci{+5.09}{3.26}{6.92} & \ci{+9.51}{7.41}{11.62} & \ci{+7.42}{4.93}{9.85} & \ci{-2.08}{-4.46}{0.26} \\
Full & \ci{+5.32}{3.47}{7.16} & \ci{+10.81}{8.60}{13.00} & \ci{+8.68}{6.16}{11.20} & \ci{-2.13}{-4.52}{0.21} \\
\bottomrule
\end{tabular*}
\end{table}

\section{Additional Privilege-Profile Evidence}
\label{app:privilege-profiles}

\subsection{Coverage and same-prefix matching}

The thinking-enabled profile uses 512 problems and four fixed no-PI trajectories per problem.
Across 18,467,995 distinct student token positions, six PI contexts plus the no-PI context produce 129,275,965 scored token--condition evaluations.
Every condition shares the span fitting all teacher contexts, capped at 1,024 completion tokens for direct-response profiles.
Both profiles require at least 512 matched tokens. This excludes 14.3\% of direct-response completions, leaving 1,756 trajectories from 466 problems.
The teacher uses the prompt below with thinking enabled in both profiles, and omitted completions are not imputed.
Correctness alignment requires both correct and incorrect scored completions, leaving 116 eligible problems for thinking-enabled profiles and 200 for direct-response profiles.
Their paired intersection contains 48 problems, so paired cross-mode estimates use a different population from the separate mode averages.

The student uses the shared block below, and the teacher appends the teacher-only block. Lines are wrapped for display.
Both use the Qwen chat template with a generation prompt. The student-generation thinking flag selects the prefix distribution, while teacher scoring keeps thinking enabled.

\par\smallskip
\noindent\begin{minipage}{\linewidth}
\textbf{Profiling prompt}\par\smallskip
\begingroup
\setlength{\fboxsep}{8pt}
\noindent\colorbox{PIShapeGroupGrey}{%
  \begin{minipage}{\dimexpr\linewidth-2\fboxsep\relax}
  \fontsize{9}{11}\selectfont
  \textbf{Shared by student and teacher}\par\medskip
  {\ttfamily\raggedright
  Solve the following math problem. Reason step by step, and put the final answer in \textbackslash boxed\{\}.
  \par\medskip
  Problem:\par
  {\color{PIShapeRoseDark}\detokenize{{question}}}\par}
  \end{minipage}%
}\par\vspace{4pt}
\noindent\colorbox{PIShapeBlush}{%
  \begin{minipage}{\dimexpr\linewidth-2\fboxsep\relax}
  \fontsize{9}{11}\selectfont
  \textbf{Teacher-only addition}\par\medskip
  {\ttfamily\raggedright
  Privileged reference available only to the teacher:\par
  <reference>\par
  {\color{PIShapeRoseDark}\detokenize{{condition_target}}}\par
  </reference>
  \par\medskip
  Use the reference to help solve or evaluate the problem, but do not mention that a reference was provided.\par}
  \end{minipage}%
}\par
\endgroup
\end{minipage}\par

\subsection{Alignment on frozen and trained-student responses}
\label{app:drift-profile}

For the trained-student profile, we generated responses to the 512 profiling problems from each seed-0 direct-response-trained student at step 50 and scored them with the same frozen teacher under all seven contexts.
Prompts and generation seeds match the pre-training procedure.
Table~\ref{tab:correctness-alignment} compares the frozen-base profiles with alignment on each trained student's own responses.
The responses and the set of problems with both correct and incorrect completions can therefore change after training; the teacher remains fixed.
Five views have much smaller estimates after training, while \FT retains about three-quarters of its initial value.

\begin{table}[!htb]
\caption{\textbf{Correctness alignment across student trajectories.}
$C_v$ in nats per token with 95\% problem-bootstrap intervals.
The teacher is thinking-enabled under both student response modes.
The final column scores each trained student's own direct-response completions at step 50.}
\label{tab:correctness-alignment}
\label{tab:drift-alignment}
\centering
\tablesetup
\begin{tabular}{@{}lrrr@{}}
\toprule
\hdrl{}{View} & \hdrr{Frozen base}{thinking enabled} & \hdrr{Frozen base}{direct response} & \hdrr{Own step-50 states}{direct response} \\
\midrule
\viewsq{PIViewAO}\AO & \ci{+0.0065}{0.0041}{0.0091} & \ci{+0.0041}{0.0028}{0.0054} & \ci{+0.0013}{0.0002}{0.0024} \\
\viewsq{PIViewGI}\GI & \ci{+0.0042}{0.0015}{0.0072} & \ci{+0.0067}{0.0045}{0.0090} & \ci{+0.0024}{0.0001}{0.0046} \\
\viewsq{PIViewKP}\KP & \ci{+0.0019}{-0.0011}{0.0051} & \ci{+0.0094}{0.0062}{0.0126} & \ci{+0.0011}{-0.0022}{0.0044} \\
\viewsq{PIViewCS}\CS & \ci{-0.0005}{-0.0033}{0.0024} & \ci{+0.0093}{0.0060}{0.0128} & \ci{-0.0003}{-0.0034}{0.0029} \\
\viewsq{PIViewSU}\SU & \ci{-0.0031}{-0.0063}{0.0001} & \ci{+0.0089}{0.0053}{0.0125} & \ci{+0.0025}{-0.0010}{0.0059} \\
\viewsq{PIViewFT}\FT & \ci{-0.0123}{-0.0173}{-0.0072} & \ci{+0.0131}{0.0087}{0.0176} & \ci{+0.0099}{0.0054}{0.0144} \\
\bottomrule
\end{tabular}
\end{table}

Figure~\ref{fig:profile-alignment-components} separates the correct- and incorrect-response shifts underlying alignment.
On average, the teacher assigns lower likelihood than the student to both kinds of response. Correctness alignment measures the difference between those mean shifts.

\begin{figure}[!htb]
\centering
\includegraphics[width=\linewidth]{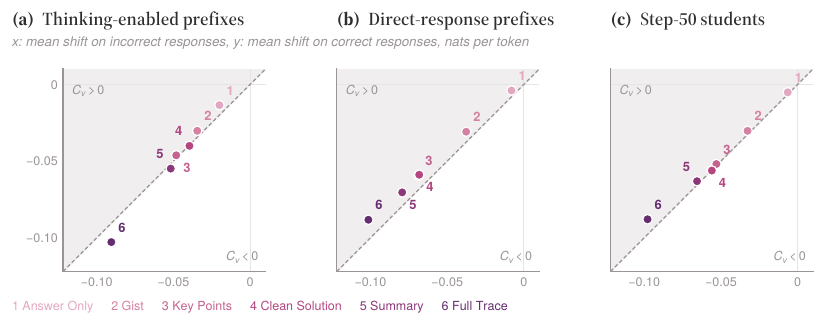}
\caption{\textbf{Correctness alignment separates two negative likelihood shifts.}
Each point compares mean per-token log-likelihood shifts on incorrect responses (horizontal axis) and correct responses (vertical axis).
The vertical distance from the equal-shift line is $C_v$.
Panels use frozen-base thinking-enabled responses, frozen-base direct responses, and each step-50 student's direct responses, averaging within problems with both outcomes.}
\label{fig:profile-alignment-components}
\end{figure}

\subsection{Sensitivity to the scored span}

Restricting the retained thinking-enabled rows to positions below 1,024 attenuates the alignment trend but leaves correction pressure negative for every view (Figure~\ref{fig:profile-matched-span}).
Thus, the full-profile comparison varies both student states and scoring horizon.
This sensitivity check uses already scored tokens rather than generating or rescoring trajectories.

\begin{figure}[!htb]
\centering
\includegraphics[width=\linewidth]{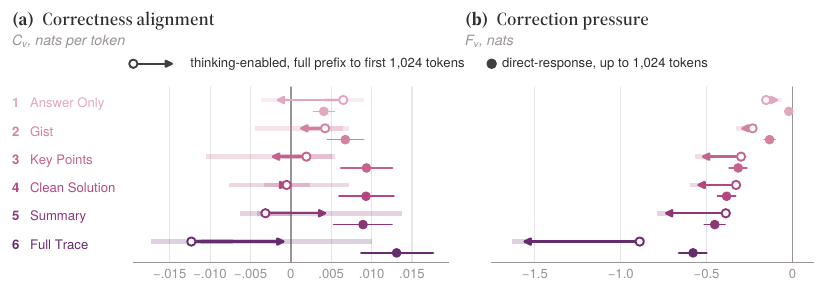}
\caption{\textbf{Profile sensitivity to the scored span.}
Arrows connect the full-prefix thinking-enabled profile to the same retained rows below position 1,024.
Filled circles show the direct-response profile, capped at 1,024 by design.
Light bands and error bars show 95\% problem-cluster bootstrap confidence intervals.
$C_v$ denotes correctness alignment and $F_v$ the shift at sampled correction markers.}
\label{fig:profile-matched-span}
\end{figure}

\subsection{Interpretation and limits of correctness alignment}
\label{app:theory}

Correctness alignment describes how privileged conditioning distinguishes correct from incorrect profiled responses.
We give its local reweighting interpretation, then show why the same aggregate diagnostics can accompany different parameter updates.

\noindent\textbf{Reweighting fixed responses.}
For a mixed-outcome problem $x$, let $\mu_x$ be the empirical distribution assigning equal weight to each profiled completion, $R(y)\in\{0,1\}$ its correctness indicator, and $h_v(y)=\Delta_v(y)$ its mean shift over the retained span from Section~\ref{sec:profile}.
Write $p_x=\E_{\mu_x}[R]\in(0,1)$, so Equation~\ref{eq:correctness-alignment} gives $C_v(x)=\E_{\mu_x}[h_v\mid R=1]-\E_{\mu_x}[h_v\mid R=0]$.
Under the exponential reweighting $\mu_{v,\epsilon}(y)\propto\mu_x(y)\exp(\epsilon h_v(y))$,
\begin{equation*}
\frac{d}{d\epsilon}\,\E_{\mu_{v,\epsilon}}[R]\,\Big|_{\epsilon=0}=\Cov_{\mu_x}(R,h_v)=p_x(1-p_x)\,C_v(x).
\end{equation*}
\emph{Proof.} Differentiating the normalized weighted mean at zero gives $\E[Rh_v]-\E[R]\E[h_v]$.
Substituting $\E[Rh_v]=p_x\E[h_v\mid R=1]$ and expanding $\E[h_v]$ by correctness yields the result. $\square$
Thus, positive $C_v(x)$ means that a sufficiently small positive reweighting raises correctness on that problem's fixed sample.
The derivative of mean correctness across problems weights $C_v(x)$ by $p_x(1-p_x)$, whereas reported $C_v$ averages problems equally.
These two quantities can rank views differently, and reweighting fixed responses is not an OPSD parameter update.

\noindent\textbf{Example: equal alignment and mean KL, opposite transfer.}
Two equally likely problems have binary one-token answers, with correct answer $1$ and student probabilities $p_\theta(1\mid x_1)=\sigmoid(\theta)$ and $p_\theta(1\mid x_2)=\sigmoid(-2\theta)$, where $\sigmoid$ is logistic and $\theta_0=0$.
Let $q_A$ and $q_B$ be frozen teacher distributions assigning correct-answer probabilities $(0.54,0.51)$ and $(0.51,0.54)$, respectively.
A same-backbone construction is $p(1\mid x_i,z)=\sigmoid(c_i\theta_0+b_i(z))$, with $c=(1,-2)$, $b_i(\varnothing)=0$, and $b_i(z_v)=\operatorname{logit}q_v(1\mid x_i)$.
Both contexts have expected teacher accuracy $0.525$ and equal mean teacher--student KL because their problem-level distributions are swapped.
For a profile containing both answers on each problem, they also have equal alignment at initialization,
\begin{equation*}
C_A=C_B=\tfrac12\Big(\log\tfrac{0.54}{0.46}+\log\tfrac{0.51}{0.49}\Big)\approx0.100,
\end{equation*}
since $C_v(x)=\operatorname{logit}q_v(1\mid x)$.
For $\Ls_v(\theta)=\tfrac12\sum_i\KL\big(q_v(\cdot\mid x_i)\,\|\,p_\theta(\cdot\mid x_i)\big)$, the gradients $G_v=\Ls_v'(0)$ are
\begin{equation*}
G_A=\tfrac12\big[-0.04+0.02\big]=-0.010,\qquad G_B=\tfrac12\big[-0.01+0.08\big]=+0.035,
\end{equation*}
while expected student accuracy $\gA(\theta)=\tfrac12[\sigmoid(\theta)+\sigmoid(-2\theta)]$ has $\gA'(0)=-1/8$.
A gradient-descent step of size $\eta$ changes accuracy by $-0.00125\,\eta+O(\eta^2)$ under $A$ and $+0.004375\,\eta+O(\eta^2)$ under $B$, giving opposite signs for sufficiently small positive $\eta$.
Every vocabulary term lies below the $0.05$ clipping threshold near initialization, so the first-step conclusion also holds for the implemented clipped divergence.
The difference comes from shared parameters: fitting $x_1$ moves $x_2$ in the opposite direction with twice the sensitivity, which the aggregate diagnostics do not capture.

\section{Intervention Evidence}
\label{app:registered-stress-tests}

The eleven direct-response control configurations comprise two other-problem references, three trace openings, four correction-marker edits, and two prompt-template changes.
We give their construction details below, followed by the loss-support intervention under thinking-enabled training.

\subsection{Reference substitutions and trace openings}
\label{app:reference-construction}

Other-problem controls replace both the reasoning body and canonical answer with a length-matched view from another problem.
Trace openings truncate genuine \FT at a token boundary to each problem's own \KP, \CS, or \SU length (means 298, 523, and 870 Qwen3 tokens across 2,112 problems).
All cuts remove the canonical answer section and later reasoning, changing content and length together. None contains a boxed answer.

Main Table~\ref{tab:mechanism-audit} gives the thinking-enabled outcomes at step 100.
Both other-problem replacements lower accuracy after within-family Holm correction. At development-selected checkpoints, only the \KP replacement's accuracy difference has an interval below zero.
Table~\ref{tab:reference-construction-modes} adds matched single-seed direct-response comparisons. All three trace openings have positive mode interactions, while the intervals for both other-problem interactions include zero.

\begin{table}[!htb]
\caption{\textbf{Reference-construction effects depend on evaluation mode.}
Step-100 Avg@4 differences, in percentage points with paired 95\% problem-cluster bootstrap intervals.
Every row uses one training seed. Trace openings are compared with seed-0 \FT, and other-problem references with seed-0 \KP or \CS, respectively.
The interaction is the direct-response contrast minus the thinking-enabled contrast on the same problems.
}
\label{tab:reference-construction-modes}
\centering
\tablesetup
\setlength{\tabcolsep}{3pt}
\begin{tabular}{@{}lrrr@{}}
\toprule
\textbf{Configuration} & \hdrr{Thinking}{enabled} & \hdrr{Direct}{response} & \textbf{Interaction} \\
\midrule
Opening, \KP length & \ci{-0.20}{-2.02}{1.69} & \ci{+9.05}{6.12}{12.04} & \ci{+9.24}{5.86}{12.63} \\
Opening, \CS length & \ci{-0.59}{-2.28}{1.11} & \ci{+11.33}{8.40}{14.32} & \ci{+11.91}{8.59}{15.30} \\
Opening, \SU length & \ci{-1.11}{-2.86}{0.65} & \ci{+7.81}{5.08}{10.55} & \ci{+8.92}{5.66}{12.24} \\
\midrule
Other-problem \KP & \ci{-2.15}{-3.97}{-0.39} & \ci{-3.91}{-6.64}{-1.17} & \ci{-1.76}{-5.01}{1.50} \\
Other-problem \CS & \ci{-1.95}{-3.78}{-0.20} & \ci{-0.39}{-3.06}{2.28} & \ci{+1.56}{-1.76}{4.82} \\
\bottomrule
\end{tabular}
\end{table}

\subsection{Correction-marker weights and prompt replacements}

Marker edits for \FT and \CS either exclude marker positions and divide by the retained-token count, or multiply their losses by $0.1$ and divide by the original supported-token count.
After the $0.05$ per-vocabulary-term clip, weighted token losses are averaged within each completion, then equally across the effective batch of 32.

Prompt swaps for \FT and \CS replace both student and teacher training wording with the original OPSD templates \citep{zhao2026opsd}.
Evaluation retains the local student prompt (Appendix~\ref{app:privilege-profiles}).
During training, both versions use one user message and a generation prompt, with student thinking off and teacher thinking on.
The replacement templates preserve the original wording and blank lines, with long lines wrapped for display.

\par\smallskip
\textbf{Original OPSD prompts used for replacement}\par\nobreak\smallskip\nobreak
\begingroup
\setlength{\fboxsep}{6pt}
\noindent\colorbox{PIShapeGroupGrey}{%
  \begin{minipage}{\dimexpr\linewidth-2\fboxsep\relax}
  \fontsize{9}{11}\selectfont
  \textbf{Student}\par\smallskip
  {\ttfamily\frenchspacing\raggedright
  Problem: {\color{PIShapeRoseDark}\detokenize{{question}}}\\[\baselineskip]
  Please reason step by step, and put your final answer within \textbackslash boxed\{\}.\par}
  \end{minipage}%
}\par\vspace{4pt}
\noindent\colorbox{PIShapeBlush}{%
  \begin{minipage}{\dimexpr\linewidth-2\fboxsep\relax}
  \fontsize{9}{11}\selectfont
  \textbf{Teacher}\par\smallskip
  {\ttfamily\frenchspacing\raggedright
  Problem: {\color{PIShapeRoseDark}\detokenize{{question}}}\\[\baselineskip]
  Here is a reference solution to this problem:\\
  === Reference Solution Begin ===\\
  {\color{PIShapeRoseDark}\detokenize{{reference text}}}\\
  === Reference Solution End ===\\[2\baselineskip]
  After reading the reference solution above, make sure you truly understand the reasoning behind each step {\rmfamily\textemdash} do not copy or paraphrase it.
  Now, using your own words and independent reasoning, derive the same final answer to the problem above.
  Think step by step, explore different approaches, and don\textquotesingle{}t be afraid to backtrack or reconsider if something doesn\textquotesingle{}t work out:\\[\baselineskip]
  Please reason step by step, and put your final answer within \textbackslash boxed\{\}.\par}
  \end{minipage}%
}\par
\endgroup

\subsection{Loss-support intervention and checkpoint sensitivity}

The one-seed \FT experiment compares the three loss windows in Section~\ref{sec:support-mismatch}.
The four 256-token distributed windows are centered near fractions $0.125$, $0.375$, $0.625$, and $0.875$ of the reasoning span before \texttt{</think>}.
All three score highest at step 25 in Table~\ref{tab:support-levels}, although development selection chooses step 50 for \textsc{Early-1K} and step 25 for the alternatives.
All paired same-step loss-window contrasts in Table~\ref{tab:mechanism-audit} have intervals that include zero.

\begin{table}[!htb]
\caption{\textbf{Loss-support checkpoint sensitivity.} Thinking-enabled \FT Avg@4 in percent, from one training seed. Paired differences and intervals are in Table~\ref{tab:mechanism-audit}.}
\label{tab:support-levels}
\centering
\tablesetup
\begin{tabular}{@{}lrrr@{}}
\toprule
\textbf{Support} & \textbf{Step 25} & \textbf{Step 50} & \textbf{Step 100} \\
\midrule
Early-1K & 78.26 & 75.13 & 72.79 \\
First-4K & 79.23 & 74.28 & 71.22 \\
Distributed-1K & 79.23 & 75.65 & 71.48 \\
\bottomrule
\end{tabular}
\end{table}

\endgroup

\end{document}